\documentclass[journal]{IEEEtran}
\usepackage{lineno}
\usepackage{setspace}
\usepackage{hyperref}
\usepackage{cite}
\usepackage{amsmath}
\usepackage{array}
\usepackage[dvipsnames]{xcolor}
\usepackage[pdftex]{graphicx}
\usepackage{tabularray}
\usepackage{booktabs}
\usepackage{rotating}
\usepackage{arydshln}
\usepackage{pifont}
\usepackage{multirow}
\usepackage{caption} 
\usepackage{subcaption}
\usepackage{float}
\newcommand{\cmark}{\ding{51}}
\begin{document}
\title{Pose Estimation from Camera Images for Underwater Inspection}

\author{Luyuan~Peng,~\IEEEmembership{Student Member,~IEEE,}
        Hari~Vishnu,~\IEEEmembership{Senior Member,~IEEE,}
        Mandar~Chitre,~\IEEEmembership{Senior Member,~IEEE,}
        Yuen~Min~Too,~\IEEEmembership{Member,~IEEE,}
        Bharath~Kalyan,~\IEEEmembership{Senior Member,~IEEE,}
        Rajat~Mishra,~\IEEEmembership{Member,~IEEE}
        
        and~Soo~Pieng~Tan
        }

\maketitle

\begin{abstract}
High-precision localization is pivotal in underwater reinspection missions. Traditional localization methods like inertial navigation systems, Doppler velocity loggers, and acoustic positioning face significant challenges and are not cost-effective for some applications. Visual localization is a cost-effective alternative in such cases, leveraging the cameras already equipped on inspection vehicles to estimate poses from images of the surrounding scene. Amongst these, machine learning-based pose estimation from images shows promise in underwater environments, performing efficient relocalization using models trained based on previously mapped scenes. We explore the efficacy of learning-based pose estimators in both clear and turbid water inspection missions, assessing the impact of image formats, model architectures and training data diversity. We innovate by employing novel view synthesis models to generate augmented training data, significantly enhancing pose estimation in unexplored regions. Moreover, we enhance localization accuracy by integrating pose estimator outputs with sensor data via an extended Kalman filter, demonstrating improved trajectory smoothness and accuracy. 
\end{abstract}

\begin{IEEEkeywords}
underwater, localization, neural networks, novel view synthesis, NeRF, sensor fusion.
\end{IEEEkeywords}

%
\IEEEpeerreviewmaketitle

\section{Introduction}
\IEEEPARstart{L}OCALIZATION plays a crucial role in underwater reinspection missions~\cite{vargas_robust_2021}. These are tasks carried out by underwater vehicles to examine the health and function of submerged structures like pipelines, offshore platforms, and ship hulls, required to ensure the safety and durability of infrastructure vital to industries like oil and gas, renewable energy, and maritime transport~\cite{bingham_robotic_2010,carreras_online_2016}. They stand apart from many underwater navigation tasks in their complexity and the precision required. Unlike general underwater navigation that involves moving from place to place, often prioritizing pathfinding and obstacle avoidance, reinspection missions demand detailed, close-range examination of often complicated underwater structures~\cite{carreras_online_2016}. As such, reinspection missions require the precise positioning and orientation of underwater vehicles to ensure thorough coverage, accurate data collection and the safety of the vehicles and the structures themselves. 

In underwater environments, the use of global positioning systems is hindered due to the rapid dissipation of electromagnetic waves in water~\cite{tan_survey_2011}. Traditionally, underwater localization has relied on inertial navigation systems (INS), Doppler velocity loggers (DVL) and acoustic positioning systems. However, these methods face significant challenges in the context of inspection missions. Acoustic navigation is often compromised by shadowing effects and multipath interference near marine structures, which can severely distort signal paths and reduce accuracy. Consequently, achieving precise acoustic navigation requires complex and costly setups~\cite{zhang_visual_2022}. Furthermore, INS and DVL, despite their widespread use, suffer from an accumulation of errors over time~\cite{zhang_visual_2022}. This limits their ability to provide the positioning accuracy required for detailed inspection of underwater structures. Although high-grade INS and DVL may be able to provide sufficient accuracy, they, too, come with high costs.  

In recent years, advancements in underwater localization have explored the use of optical sensors, such as cameras~\cite{kuutti_survey_2018}. Some of these approaches necessitate the deployment of active markers~\cite{kuutti_survey_2018, buchan_low-cost_2017} or elaborate setups by divers~\cite{gomez_chavez_robust_2018,chiarella_gesture-based_2015}, adding complexity and expense. In contrast, visual localization methods—estimating camera poses from images of the surrounding scene—present a more cost-effective solution. Since inspection vehicles typically come equipped with cameras, visual-based localization can be implemented without the need for extra hardware. Moreover, visual-based localization methods, such as simultaneous localization and mapping (SLAM)~\cite{zhang_visual_2022}, visual odometry~\cite{teixeira_deep_2020} and visual relocalization~\cite{kendall_posenet_2015, peng_improved_2024}, have shown promise in navigating terrestrial and underwater environments. 

Underwater reinspection missions typically involve the vehicle returning to the same sites for routine monitoring, assessment and/or maintenance. In this sense, these missions have another difference from normal underwater navigation tasks in that they have prior information of the scene or environment available, i.e., the environment is ``known" to some degree after the first mission. We can use this available prior information to perform relocalization. This approach can be made effective if in the initial mapping run, we collect positioning information as accurately as possible using precise (and typically expensive and complex) positioning infrastructure such as ultra-short baseline acoustic positioning to characterize the environment. Using these data collected, visual relocalization methods can directly estimate poses from camera images in the following runs, significantly reducing the cost and complexity of reinspection. While SLAM and visual odometry are effective for general navigation, they do not utilize the additional prior information available in reinspection missions. In contrast, visual relocalization uses prior information and thus allows us to use more affordable vehicles and setups for localization in subsequent reinspection missions, significantly simplifying operations.

Visual relocalization techniques are categorized into feature-based methods such as Active Search~\cite{sattler_efficient_2017}, and deep-learning methods like PoseNet~\cite{kendall_posenet_2015}. Active search achieves image-based localization by systematically identifying and matching 2D features in query images with 3D points in a scene model. PoseNet is a deep learning model that utilizes a pretrained convolutional neural network (CNN) to estimate the 6-degree-of-freedom (6-DOF) poses of a camera directly from images. This approach simplifies the camera relocalization problem by bypassing the traditional feature extraction and matching steps, instead relying on the CNN to learn and estimate the camera's position and orientation within a previously mapped environment directly from the image data.  

While Active Search achieves state-of-the-art results in outdoor terrestrial scenes, its effectiveness and robustness are reduced in environments with sparse features or where textures are obscured, conditions common in underwater settings due to limited visibility~\cite{kendall_posenet_2015}. Additionally, Active Search is more computationally expensive compared to PoseNet~\cite{shavit_introduction_nodate}. Hence, we focus on neural-network-based methods inspired by PoseNet to estimate poses from underwater images. Previous research has demonstrated PoseNet's efficacy in conducting inspection tasks within tanks with toy structures and simulated underwater environments~\cite{nielsen_evaluation_2019,peng_regressing_2022, peng_improved_2024}. However, the performance of machine learning-based pose estimators with realistic structures and in at-sea environments has not been thoroughly investigated.

The performance of learning-based pose estimators depends heavily on the diversity of the training data. However, in underwater environments, collecting comprehensive training data is expensive. We propose to use Novel View Synthesis (NVS) models to render augmented training data. Recent advancement in NVS models, such as Neural Radiance Fields (NeRF)~\cite{mildenhall_nerf_2020} and 3D Gaussian Splatting (3DGS)~\cite{kerbl_3d_2023}, can synthesize photorealistic views of complex 3D scenes from a sparse set of input views by optimizing an underlying continuous volumetric scene function. When provided with a camera pose, NVS models utilize classical volumetric rendering techniques to project synthesized colors and densities into an image~\cite{mildenhall_nerf_2020}. Using a trained NVS model, we can render images from any viewpoint within the boundary, allowing us to bypass the need for extensive physical data collection. We can then use these rendered images to augment our training data.

In this paper, our contributions are as follows:
\begin{enumerate}
\item We examine the performance of neural-network based pose estimators with different configurations in inspection missions in confined waters. We investigate the effects of different parameters, such as using RGB information versus grayscale, on the performance. We present the dataset collected, methods employed and results obtained in Section \hyperref[sec:2]{II}.
\item We propose a new loss function, $d$-loss, incorporating the geometry of the inspection missions for training the pose estimators. The $d$-loss provides interpretability, and improves computation efficiency and estimation performance. We present the method and results in Section \hyperref[sec:2]{II}.
\item We utilize underwater 3D NVS techniques to generate augmented training data. We demonstrate the performance improvement due to this in Section ~\hyperref[sec:3]{III}.
\item We enhance the localization performance by integrating our pose estimation model with data from additional sensors, such as altimeters and compasses. We use an extended Kalman filter (EKF) for tracking and fusion. In Section \hyperref[sec:4]{IV}, we present these methods and results showing improved robustness and accuracy of this approach.
\item We evaluate the performance of our proposed methods in at-sea environments. We present these results and discuss the overall performance of the entire pipeline in Section \hyperref[sec:5]{V}.
\end{enumerate}
Finally, we conclude this paper in Section \hyperref[sec:6]{VI}. 

\section{Pose Estimation}
\label{sec:2}

Nielsen et al.~\cite{nielsen_evaluation_2019} evaluated the performance of PoseNet in a small tank, inspecting a subsea connector attached to a metal stick. In our previous work, we assessed the performance of various pretrained CNNs as pose estimators in a simulated underwater environment inspecting a subsea pipe~\cite{peng_regressing_2022}. In this section, we evaluate the performance of visual localization using two neural-network model architectures inspired from PoseNet~\cite{kendall_posenet_2015}.
The data for training and testing were collected from an artificial ocean basin at the Technology Center for Offshore and Marine, Singapore (TCOMS)~\cite{TCOMS}. The originally presented PoseNet \cite{kendall_posenet_2015} works on RGB images. Here, we also evaluate the visual localization performance using grayscale images instead of RGB images to determine if similar accuracy can be achieved with higher efficiency, based on the intuition that underwater images typically have limited color information. Finally, we investigate the models' capability for (1) estimating pose on test images from the same dataset (i.e., capability to interpolate within same dataset), and (2) their capability to generalize to datasets outside that used for training, by using data from different runs for training and testing, which have different paths and conditions during acquisition.

\subsection{Methods}
\subsubsection{Architecture}
The objective of PoseNet is to estimate a 6-DOF pose from a single monocular RGB image given as input to a neural network. The pose consists of the position (in 3D coordinates,  $x$-$y$-$z$) and the orientation, which is represented in terms of a quaternion. Thus, the model outputs a 7-dimensional (7D) estimated pose vector $\mathbf{y} = [\hat{\mathbf{p}}, \hat{\mathbf{q}}]$ containing a position vector estimate $\hat{\mathbf{p}}$ and an orientation vector estimate $\hat{\mathbf{q}}$, where $\hat{ }$ represents an estimate. 

\begin{figure*}[t]
\centerline{\includegraphics[width=\textwidth]{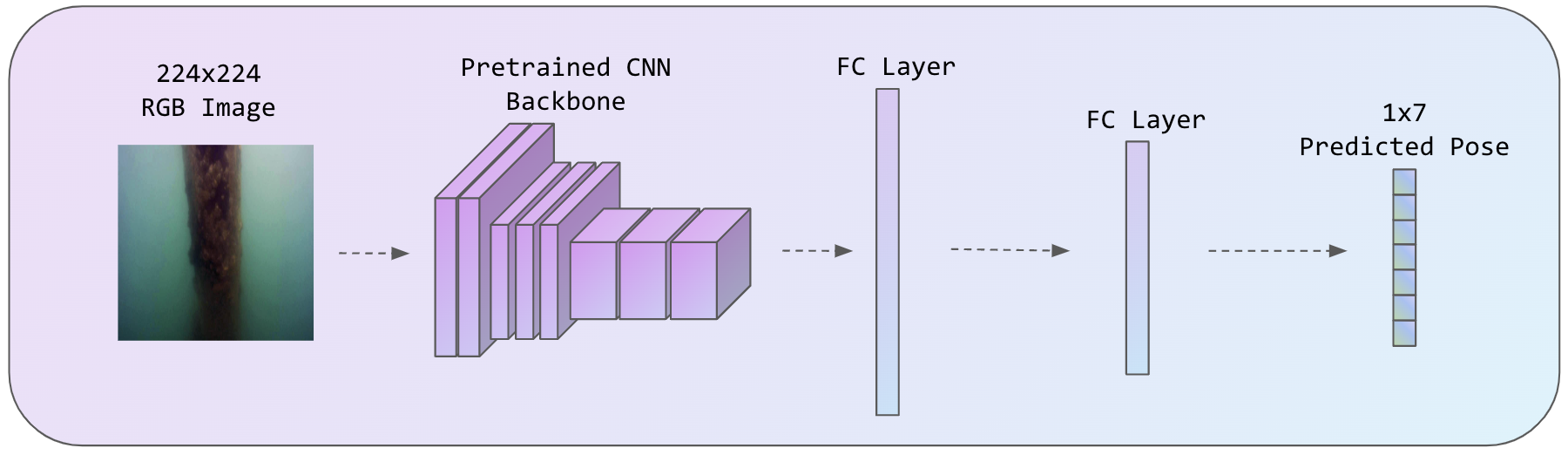}}
\captionsetup{justification=centering}
\caption{Overview of the CNN-based architecture for visual localization.}
\label{fig:net}
\end{figure*}

\begin{figure*}[t]
\centerline{\includegraphics[width=\textwidth]{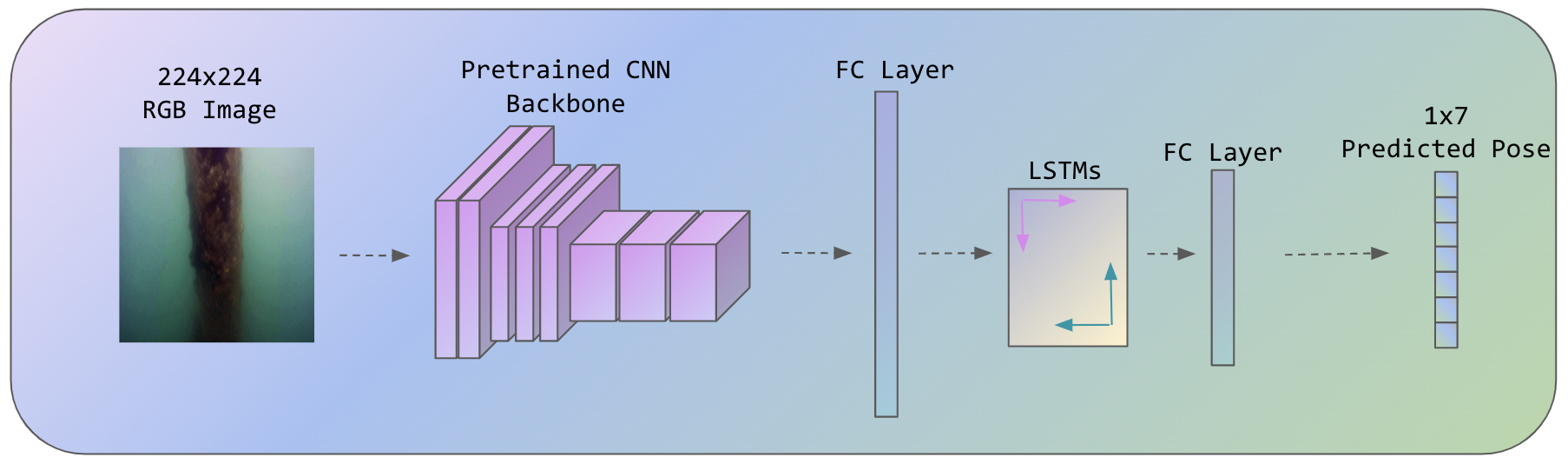}}
\caption{Overview of the CNN+LSTM-based architecture for visual localization}
\label{fig:lstm}
\end{figure*}

The PoseNet model originally presented by Kendall et al~\cite{kendall_posenet_2015} was a CNN, a modified version of the GoogLeNet architecture~\cite{szegedy_going_2014} pretrained on the ImageNet dataset~\cite{imagenet_cvpr09}, with the softmax classifiers changed to affine regressors, and another fully connected (FC) layer of feature size 2048 inserted before the final regressor. However, regressing a 7D pose vector from a high dimensional output of the FC layer is not optimal~\cite{walch_image-based_nodate}. A later work ~\cite{walch_image-based_nodate} aimed to tackle this by modifying PoseNet by reshaping the FC layer of size 2048 to a 32 $\times$ 64 matrix and applying four long-short-term-memory networks (LSTMs) to perform structured dimensionality reduction. This algorithm, which we refer to as CNN+LSTM, showed a performance improvement compared to PoseNet in terrestrial environments~\cite{walch_image-based_nodate}, and also in an underwater tank environment~\cite{peng_improved_2024}. We implement and evaluate both model architectures -- the CNN (shown in Fig.~\ref{fig:net}) and the CNN+LSTM (shown in Fig.~\ref{fig:lstm}). Additionally, we assess the performance of these using a pretrained ResNet50~\cite{he_deep_2015} as the backbone. 

\subsubsection{Loss Function}
Kendall et al~\cite{kendall_posenet_2015} used a composite loss function that is a weighted sum of the (1) L2 loss $\mathcal{L}_\mathbf{p}$ between the predicted positions and the true positions, and the (2) L2 loss $\mathcal{L}_\mathbf{q}$ between the predicted quaternions and the true quaternions: 
\begin{equation}\label{eq:posenet_loss_func}
    \mathcal{L} = \mathcal{L}_\mathbf{p}+ \beta \mathcal{L}_\mathbf{q}, 
\end{equation}
where $\mathcal{L}_\mathbf{p} = ||\mathbf{p} - \hat{\mathbf{p}}||_2$ and $\mathcal{L}_\mathbf{q} = ||\mathbf{q} - \hat{\mathbf{q}}/\|\hat{\mathbf{q}}\|||_2$, and $\mathbf{p}$ and $\mathbf{q}$ represent the true pose. $\beta$ is a free parameter that determines the trade-off between the desired accuracy in translation and orientation. In PoseNet and CNN+LSTM, the value of $\beta$ is fine-tuned using a grid search to ensure the expected value of position and orientation errors are approximately equal, which the authors suggest lead to overall optimal performance. We refer to this loss function as the $\beta$-loss.  

We argue that the $\beta$-loss is not the optimal approach to our problem, due to three reasons. Firstly, we argue that optimal performance is not necessarily achieved when position and orientation errors are roughly equal. Instead, the performance criteria and loss should incorporate geometry and physics relevant to the inspection task at hand. Secondly, the L2 loss between the predicted and true quaternions does not directly translate to an orientation error interpretable in degrees or radians, and thus, it does not accurately reflect the geometric distance between the predicted and true orientations. Thirdly, searching for the optimal $\beta$ value often involves extensive computational resources. This search can become a significant bottleneck, especially in scenarios where training needs to be done fast.

To overcome these shortcomings, we propose a new loss function more relevant to our problem, the $d$-loss, to improve the training effectiveness, interpretability and efficiency. The $d$-loss is defined as:
\begin{equation}
    \mathcal{L} = \mathcal{L}_\mathbf{p}+ d \mathcal{L}_\theta. \label{eqn:my_loss_func}
\end{equation}
Note that we have replaced the quaternion loss in (\ref{eq:posenet_loss_func}) with a loss based on the Eulerian angular difference, $\mathcal{L}_\theta$, which is calculated as follows. We first determine the rotation between the estimated and ground truth quaternions through quaternion multiplication, $\mathbf{\Delta q} = \mathbf{q} \left(\hat{\mathbf{q}}/\|\hat{\mathbf{q}}\|\right)^{*}$, where $^{*}$ denotes the conjugate of the quaternion. $\mathbf{\Delta q}$ is a unit quaternion which can be expressed as $(r, \vec{v})$ where $r$ is the scalar part of the quaternion, and $\vec{v}$ is the vector part. $r$ is related to a spatial rotation around a fixed point of $\mathcal{L}_\theta$ radians about a unit axis by $r =\cos(\mathcal{L}_\theta/2)$~\cite{bernardes_quaternion_2022}, thus $\mathcal{L}_\theta = 2\arccos(r)$. We approximate $\mathcal{L}_\theta \approx \frac{\pi}{2} (1 - r)$, using a Taylor series approximation. The Eulerian angular difference loss provides a more intuitive and direct measure of orientation error. 

Additionally, we replace the hyperparameter weight factor $\beta$ in (\ref{eq:posenet_loss_func}) which required tuning, with the average distance $d$ between the camera and the object of interest. The intuition here is that this factor translates the rotational error to an equivalent ``average" translational error (attributed to the orientation difference). Thus, the overall loss can be interpreted as the ``total positional error" in meters, including contributions from translational and orientation error components. 

\begin{figure}[]
\centerline{\includegraphics[width=0.5\textwidth]{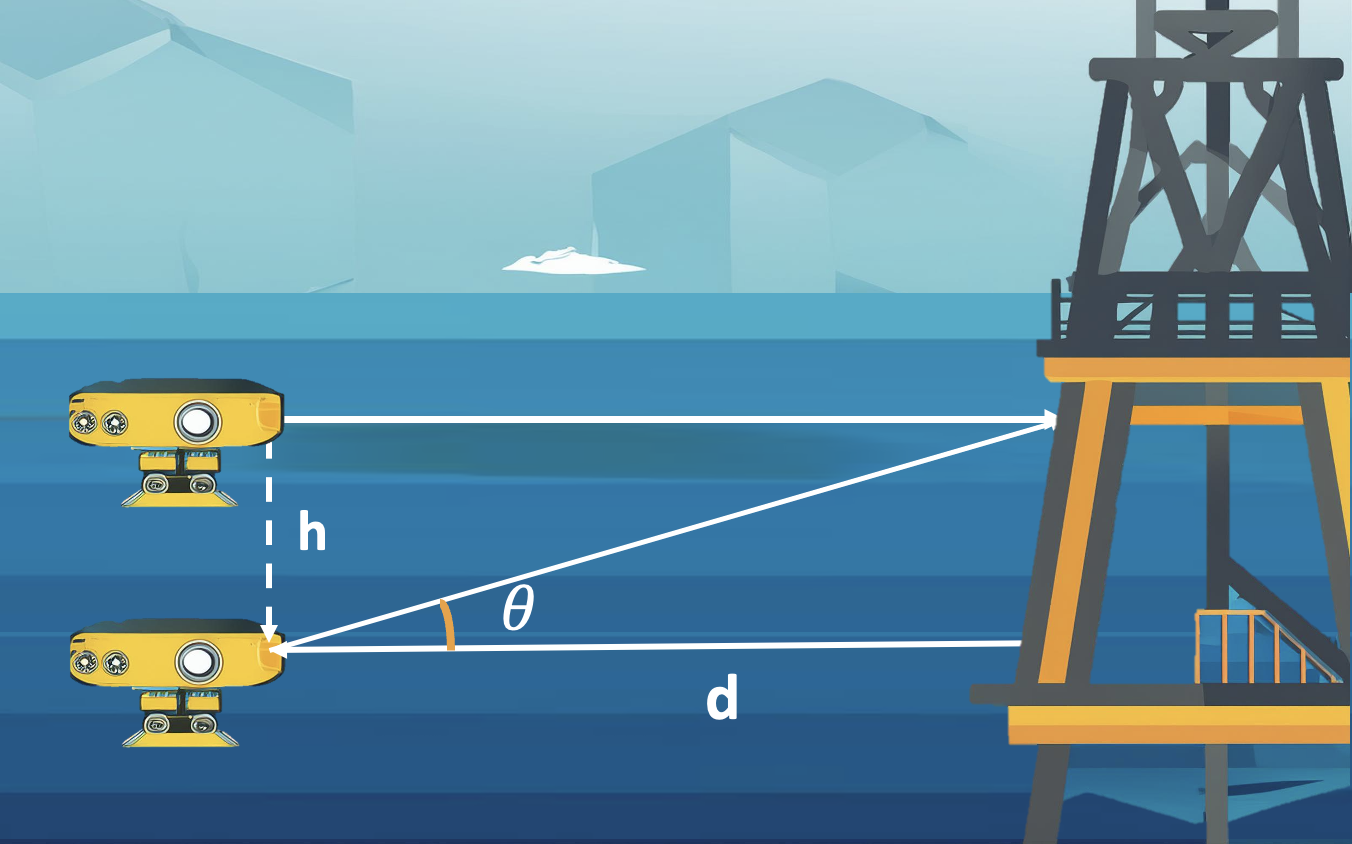}}
\caption{Schematic showing the interpretation of the orientation error in terms of equivalent translational error.}
\label{fig:angular_translation}
\end{figure}

The translation between rotational error and the ``average" translational error is described as follows. As illustrated in the example in Fig.~\ref{fig:angular_translation}, if the camera has a pitch orientation error $\mathcal{L}_\theta$ of $\theta$, the point it observes on the structure remains roughly the same as if the camera had an equivalent translational error $\mathcal{L}_\mathbf{p}$ of $h$ (i.e., moves up by $h$) for small values of $h$ and $\theta$. Based on the geometry, equivalent translational error can be expressed in terms of orientation error $\mathcal{L}_\theta$ and the average horizontal range between the camera and the structure as:
\begin{equation}
 \mathcal{L}_\mathbf{p} = d {\tan(\mathcal{L}_\theta)}.
\end{equation}
Assuming the case when the rotational error is small, we approximate $\tan(\mathcal{L}_\theta) \approx \mathcal{L}_\theta$. Thus, we obtain:
\begin{equation}
    \mathcal{L}_\mathbf{p} \approx d \mathcal{L}_\theta.
\end{equation}

This modified loss function (Eq.~\ref{eqn:my_loss_func}) leverages the inherent geometric relationship between positional and rotational errors in inspection missions, reducing computational complexity in finding the optimal $\beta$ as well as provides a more intuitive and interpretable overall loss function which has a physical meaning and represents the pose error in terms of meters. 

\subsubsection{Implementation}
To evaluate the effectiveness of deeper backbones, additional LSTM layers, the proposed $d$-loss, and the color information in images, we tested multiple configurations of the two visual localization network architectures. The details of these configurations are summarized in Table~\ref{tab:configurations}.
\begin{table}[ht]
    \centering
    \caption{Description of configurations}
    \begin{tabular}{ccccc}
    \toprule
    \textbf{ID} & \textbf{Architecture} & \textbf{Backbone} & \textbf{Loss} & \textbf{Color} \\
    \midrule
    C1 & CNN & GoogLeNet & $\beta$-loss & RGB \\
    C2 & CNN & GoogLeNet & $d$-loss & Grayscale \\
    C3 & CNN & GoogLeNet & $d$-loss & RGB \\
    C4 & CNN & ResNet50 & $d$-loss & RGB \\
    C5 & CNN+LSTM & GoogLeNet & $d$-loss & RGB \\
    C6 & CNN+LSTM & ResNet50 & $d$-loss & RGB \\
    \bottomrule
    \end{tabular}
    \label{tab:configurations}
    \end{table}

To adapt the pretrained GoogLeNet architecture to operate on grayscale images, we modified the network's first convolutional layer by adjusting it to accept single-channel (grayscale) inputs instead of the original three-channel (RGB) inputs. This adjustment was achieved by reducing the number of input channels from three to one. The single-channel input layer was initialized by summing the weights across the RGB channels in the original network to utilize as much prior information as possible. Following this, the layer was fine-tuned via training. 

During both training and testing for all configurations, we rescaled input images directly into a 224$\times$224 pixels input, deviating from PoseNet's approach of resizing the images to 256$\times$256 before cropping into 224$\times$224. This adjustment was made to minimize the loss of image information, a concern particularly acute in underwater images where available information is inherently more limited compared to terrestrial settings. To speed up training, we normalized the images against the ImageNet dataset's mean and standard deviation. Additionally, poses are normalized to lie within the range [-1, 1].

We used the PyTorch deep learning framework to implement and train the models. The experiments were conducted using an RTX 6000 Ada GPU. For training, we used the stochastic gradient descent optimizer for configurations C1, C2, and C3. For the remaining configurations, we used the Adam optimizer. A batch size of 32 was used. Hyperparameters, including the learning rate, weight decay, and $\beta$ for C1, were tuned using grid search strategy over a predefined set of values. The best set of hyperparameters was selected based on validation performance. Training continued until early stopping was triggered.

\subsection{Testing in Controlled Environment}
\begin{figure}[]
    \centerline{\includegraphics[width=0.5\textwidth]{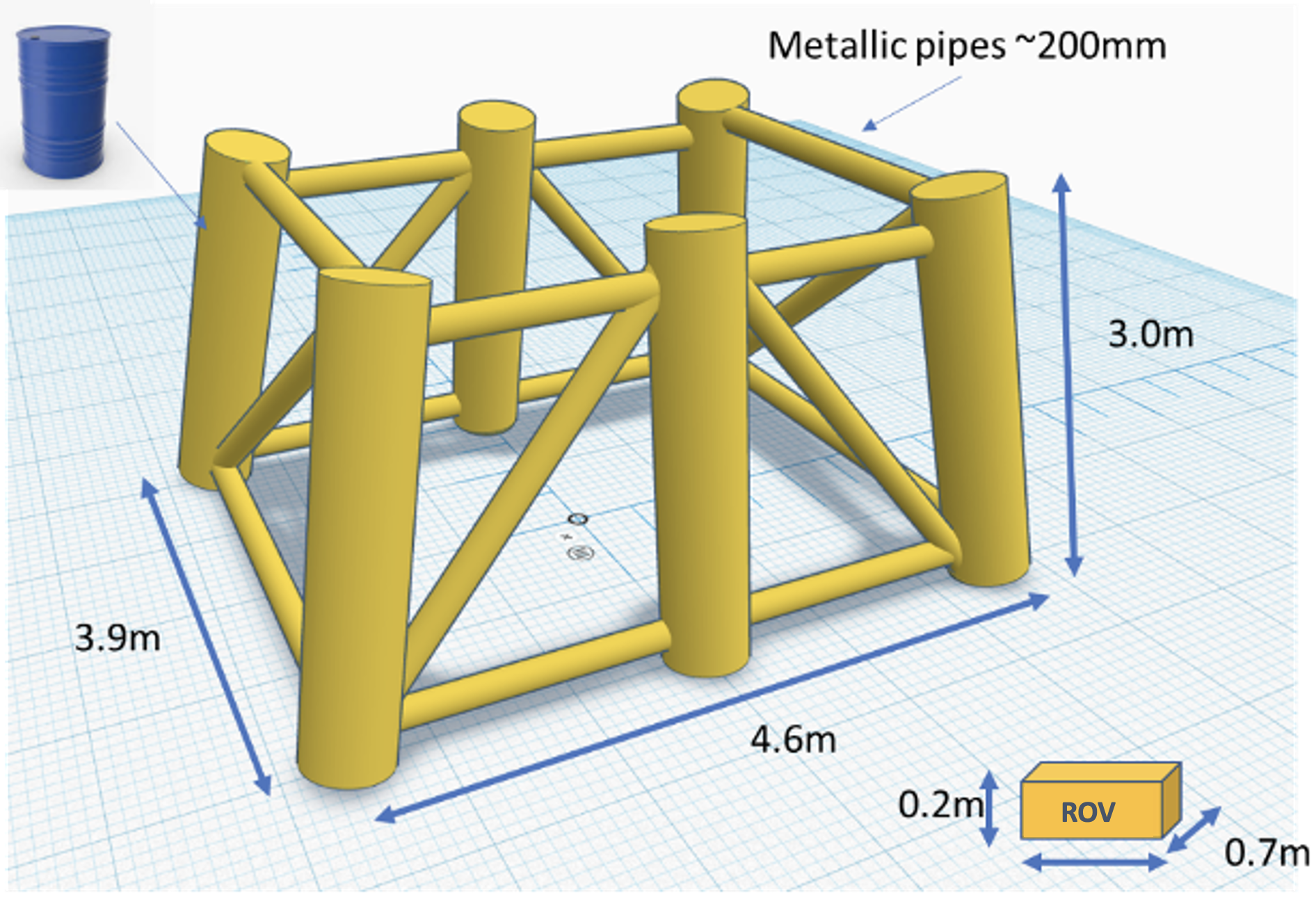}}
    \caption{Schematic of the structure surveyed in the TCOMS facility.}
    \label{fig:tcoms_struct}
    \end{figure}

\begin{figure}[]
    \centerline{\includegraphics[width=0.5\textwidth]{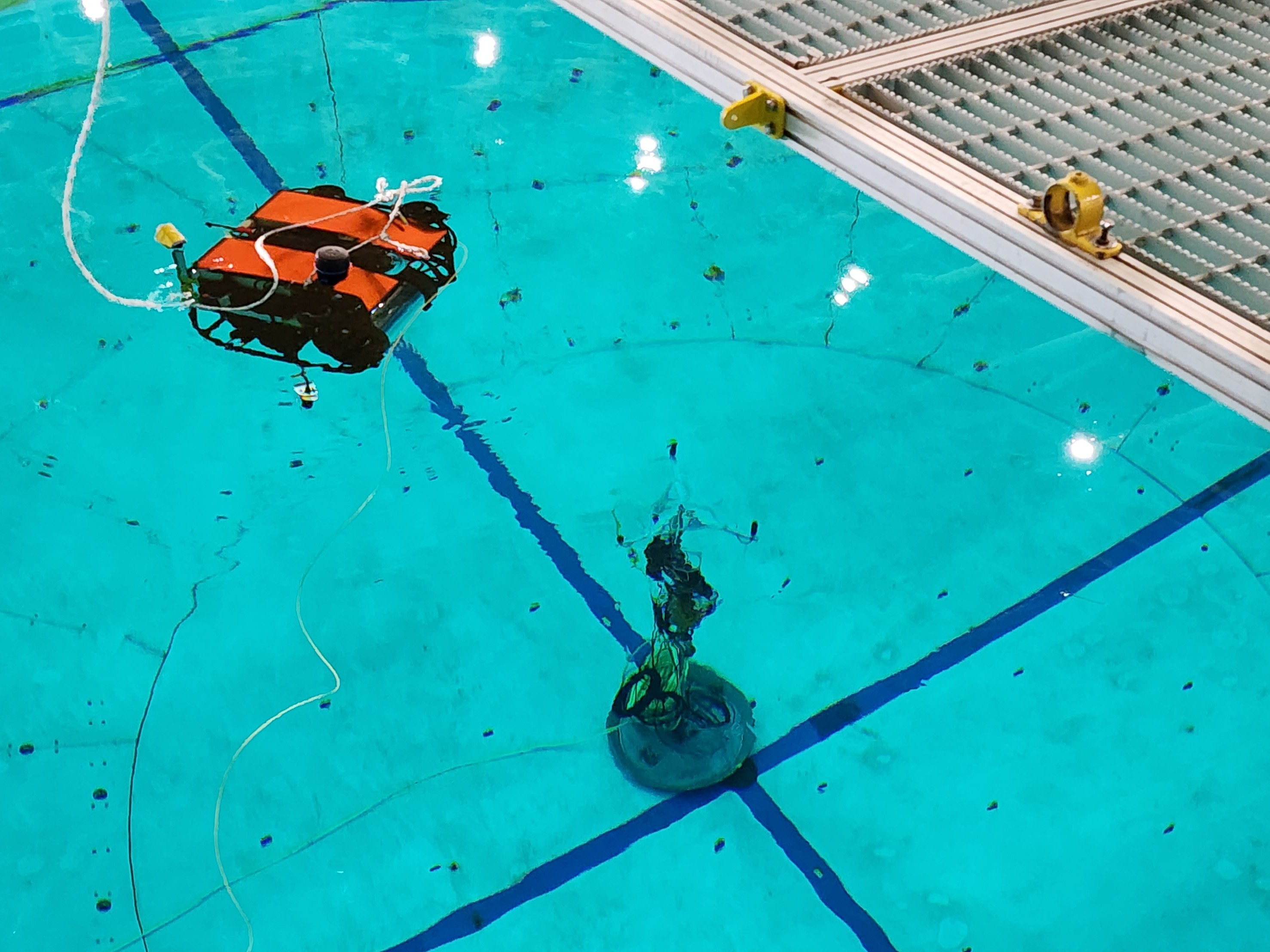}}
    \caption{The USBL setup at TCOMS to estimate the location of the ROV (shown).}
    \label{fig:tcoms_usbl}
    \end{figure}

The artificial ocean basin at TCOMS is an indoor pool measuring 60~m $\times$ 48~m $\times$ 12~m. As illustrated in Fig.~\ref{fig:tcoms_struct}, a structure was placed in the basin, which consisted of six piles interconnected by metallic pipes, with each pile comprising three metallic oil barrels. The overall dimensions of the structure were approximately 3.9~m $\times$ 4.6~m $\times$ 3.0~m. The whole structure was yellow in color. 
We used a customized remotely operated vehicle (ROV) which is equipped with a monocular camera for collecting RGB image data, a compass for collecting orientation information, and an altimeter for collecting altitude information. We placed a high frequency acoustic modem with four receivers near the operating region (as shown in Fig.\ref{fig:tcoms_usbl}) to estimate the position of the ROV using ultra-short baseline (USBL) positioning. We operated the ROV to inspect the piles in a lawnmower path. 

We executed three trials within the environment at different depths to gather data while the ROV surveyed the structure, with each trial featuring a roughly similar trajectory. We refer to these trials as D1, D2 and D3, corresponding to respective average depth levels -1.5~m, -3~m and -4~m. The details of the datasets are described in Table~\ref{tab:datasets}. 

\begin{table}[ht]
    \centering
    \caption{Description of datasets}
    \begin{tabular}{ccc}
    \toprule
    \textbf{ID} & \textbf{Dataset Name} & \textbf{Dataset Size} \\
    \midrule
    D1 & Clear Water-Deep & 2165 \\
    D2 & Clear Water-Shallow & 2956 \\
    D3 & Clear Water-Mid & 933\\
    D4 & Clear Water-NVS & 4193\\
    D5 & Sea Water-1 & 2360 \\
    D6 & Sea Water-2 & 735 \\
    D7 & Sea Water-NVS & 18918 \\
    \bottomrule
    \end{tabular}
    \label{tab:datasets}
    \end{table}

The sensor data from the vehicle was captured using ROS (Robot Operating System), and we sampled the dataset at a frequency of 5~Hz from each recorded data file. We synchronized the sampled data with the USBL position information based on timestamp and interpolate the position data when necessary. For our ground truth, we utilized the x and y coordinates from the USBL position estimates, the z coordinate from the altimeter, and the orientation data from the compass.

We use D1 as the primary dataset to evaluate the models' capability of interpolation. We randomly select 60\% points from the data for training, 20\% for validation and 20\% for testing. We further assess the models' ability to generalize to new depths by employing D1 as the training dataset and D3 for validation and testing. Additionally, we investigate the impact of incorporating data from diverse depths on the models’ generalization performance by using D1 and D2 together as the training data and D3 as the validation and test data. 

\subsection{Results \& Discussions}
\subsubsection{Model performance}
We present the performance of different configurations in Table~\ref{tab:config_perf}. The benchmark for our evaluation is the performance of C1. 

\begin{table}[ht]
    \centering
    \caption{Performance of all configurations trained and tested on dataset D1. $\mathcal{L}_\mathbf{p}$ and $\mathcal{L}_\theta$ tabulated are the median of estimates across the test data. $\mathcal{L}$ was calculated using Eq.~\ref{eqn:my_loss_func} with $d = 3$~m. The best performance for each metric is highlighted in bold.}
    \begin{tabular}{ccccc}
    \toprule
    \textbf{ID} & \textbf{$\mathcal{L}$ (m)} & \textbf{$\mathcal{L}_\mathbf{p}$ (m)} & \textbf{$\mathcal{L}_\theta$ (\textdegree)} & \textbf{Inference} \\
    \textbf{} &  &  &  & \textbf{time (ms)} \\
    \midrule
    C1 & 2.41 & 2.36 & \textbf{0.86} & 2.20 \\
    C2 & 0.61 & 0.53 & 1.50 & 1.65 \\
    C3 & 0.41 & 0.36 & 0.99 & 1.62 \\
    C4 & 0.34 & 0.29 & 0.88 & 1.16 \\
    C5 & 0.30 & 0.22 & 1.51 & 0.78 \\
    C6 & \textbf{0.19} & \textbf{0.12} & 1.34 & \textbf{0.77} \\
    \bottomrule
    \end{tabular}
    \label{tab:config_perf}
\end{table}

We observe the following: 

1. Comparing the performance of C3 against C1, our results demonstrate that training with our proposed $d$-loss significantly enhances model performance, especially in terms of the overall performance metric $\mathcal{L}$. 

\begin{figure*}[t]
    \centering
    \makebox[0.24\textwidth]{\textbf{Camera image}} 
    \hfill
    \makebox[0.72\textwidth]{\textbf{Rendered images}}
    \hfill
    \vspace{5pt} 

    \fbox{ 
    \begin{subfigure}[t]{0.224\textwidth}
        \includegraphics[width=\textwidth]{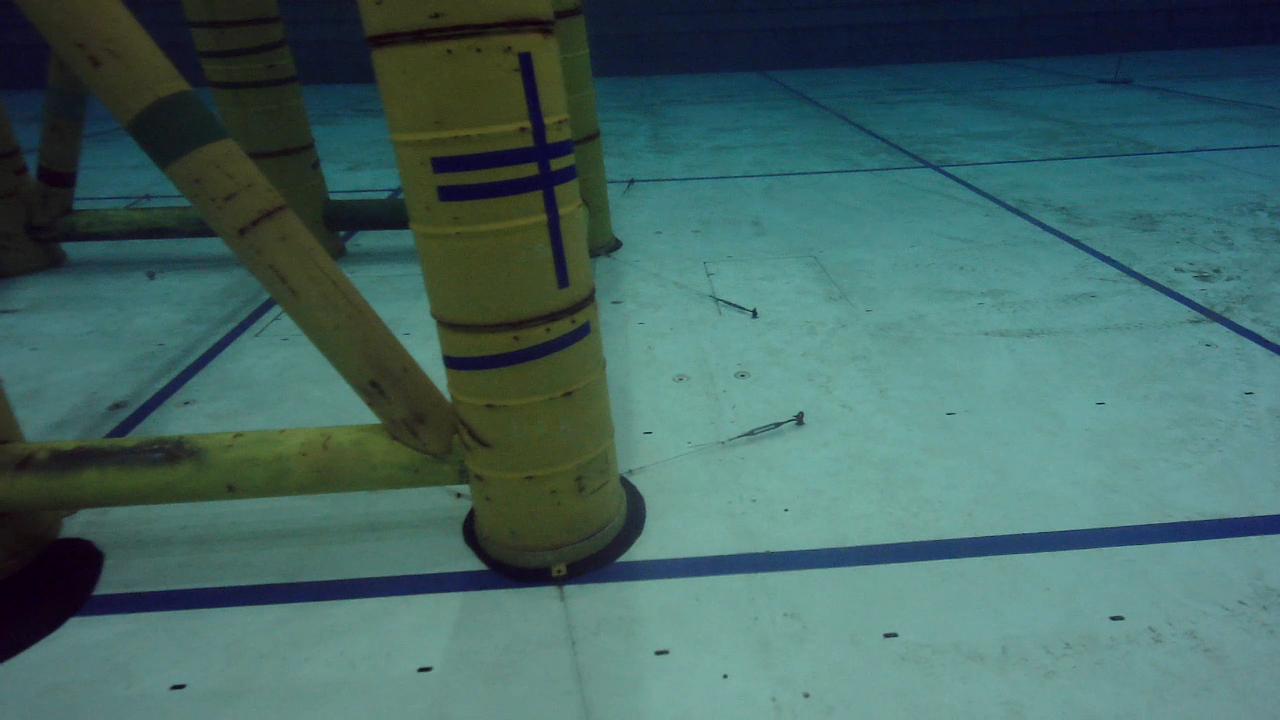}
    \end{subfigure}
    }
    \hfill
    \fbox{ 
    \begin{minipage}[t]{0.7\textwidth}
        \begin{subfigure}[t]{0.32\textwidth}
            \includegraphics[width=\textwidth]{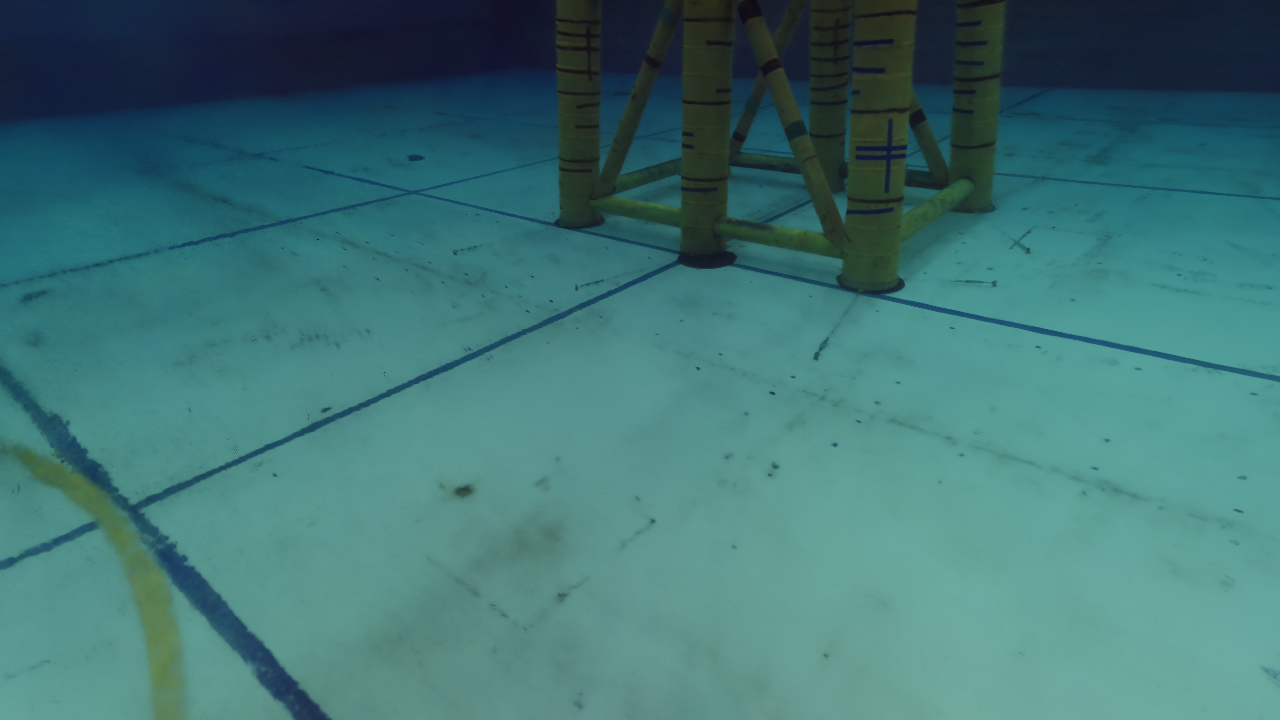}
        \end{subfigure}
        \hfill
        \begin{subfigure}[t]{0.32\textwidth}
            \includegraphics[width=\textwidth]{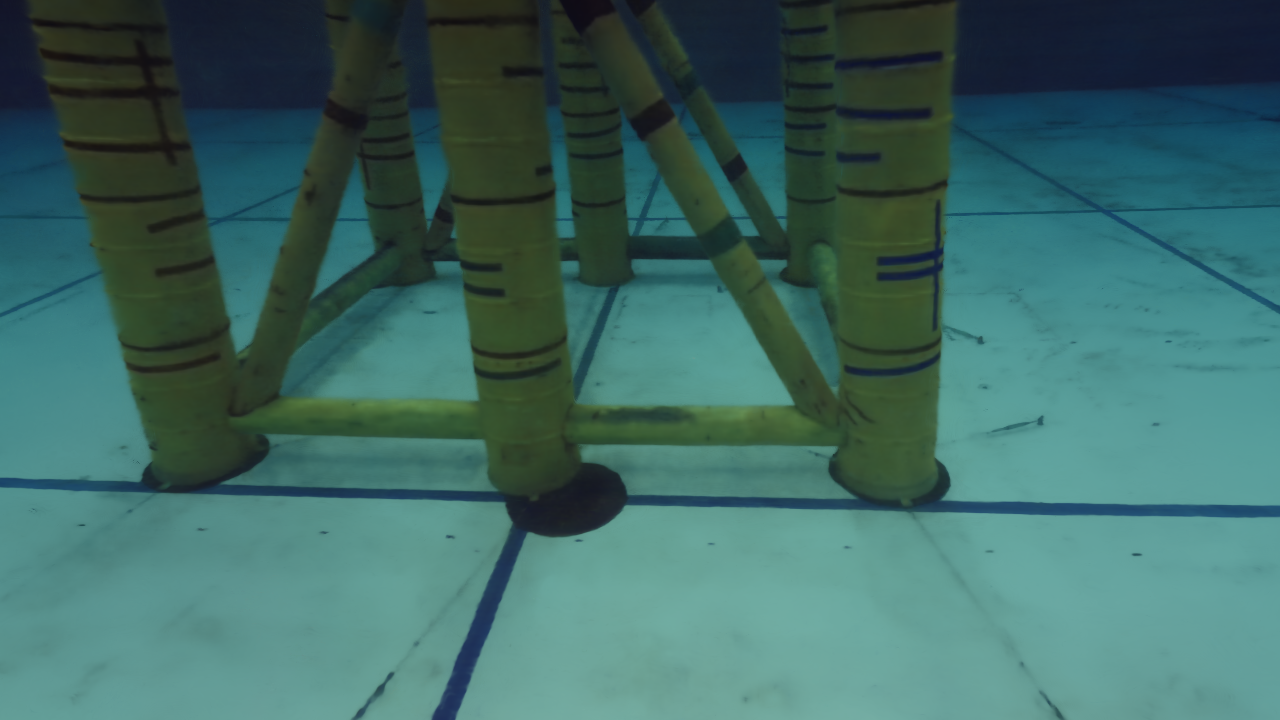}
        \end{subfigure}
        \hfill
        \begin{subfigure}[t]{0.32\textwidth}
            \includegraphics[width=\textwidth]{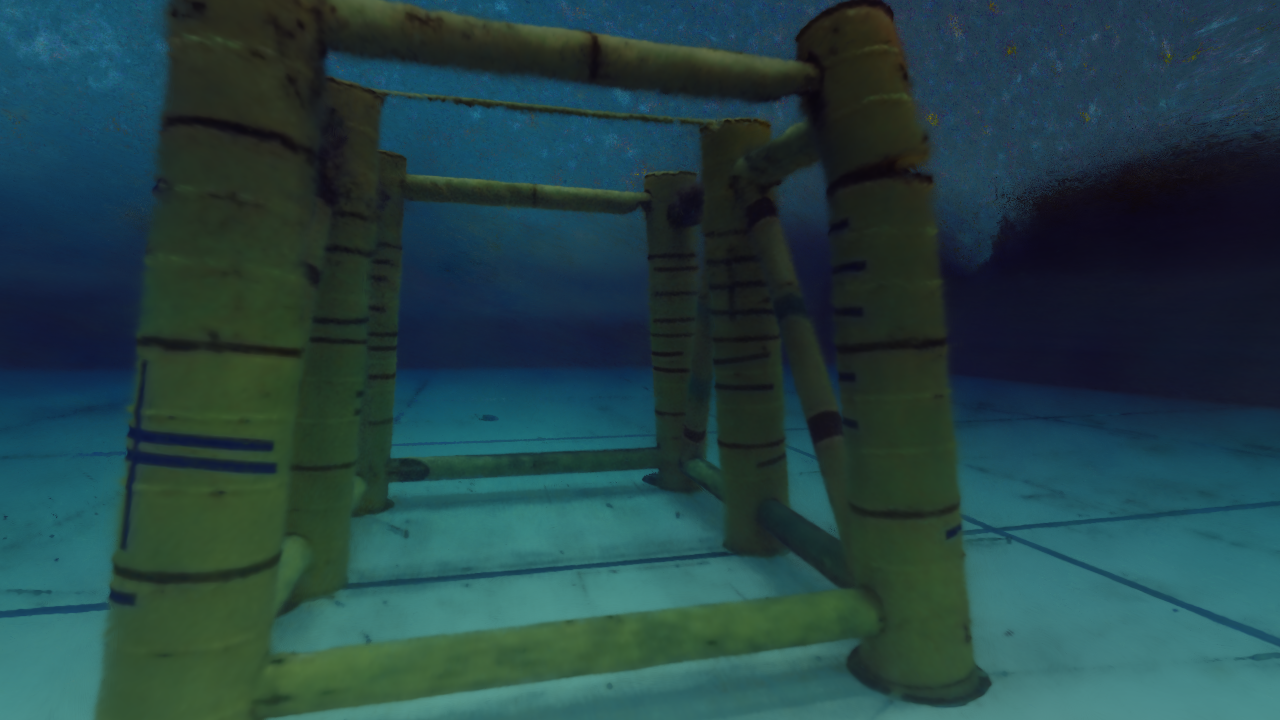}
        \end{subfigure}
    \end{minipage}
    }
    \caption{Camera images and NVS rendered images in the controlled experiment in TCOMS. The rendered images produce photorealistic views of the structure but exhibit discrepancies in brightness. Some of the rendered views have artifacts in the background as shown in the right-most image.}
    \label{fig:tcoms-images}
\end{figure*}

2. Comparing the performance of C2 against C3, it can be observed that using grayscale images shows significantly worse performance and too little an improvement in inference time, contrary to our initial expectation. The worse performance of grayscale images can be attributed to the fact that since D1 was collected in a non-turbid fresh water environment, the color information in the underwater images is not as limited as one might anticipate in an image taken in a sea environment. As shown in Fig.~\ref{fig:tcoms-images}, the underwater RGB images in D1 retain valuable color information that may provide distinguishing features in these environments. Thus, the grayscale images have much less information than RGB images and thus lead to poorer performance. The lack of improvement in inference time is due to the fact that we only reduce the number of channels in the first CNN layer of the pretrained model, resulting in a minimal reduction in computational load. To achieve more substantial computational savings, the entire model architecture would need to be better streamlined for grayscale images, not just the initial layer.
 
3. Comparing the performance of C6 to C5 and C4 to C3 shows that using ResNet50, a deeper network, as the backbone, improves performance for both CNN and CNN+LSTM. 

4. Comparing the performance of C6 to C4 and C5 to C3 shows that the CNN+LSTM architecture consistently outperforms the CNN architecture. 

Among all the configurations, C6, which uses the CNN+LSTM architecture with the ResNet50 backbone and is trained using the proposed $d$-loss, performs the best, achieving 0.12~m of positional accuracy and 1.34\textdegree  of orientation accuracy with an inference time of 0.77~ms. 

\subsubsection{Generalization performance}

We test the performance of generalization using the model with the best configuration, C6. We first trained the model on D1 and tested on D3. A significant performance degradation is observed, as shown in the first row of Table~\ref{tab:generalization}. This is on expected lines because the test data is sampled from a different distribution than the training data with possibly different paths and conditions, and deep-learning models often fail to extrapolate beyond the bounds of the training data. 

To address this issue, we evaluate the use of a larger and more diverse training dataset, by expanding the training data to include both D1 and D2. This augmentation introduces a wider distribution of data, notably enhancing the diversity in depth information. This leads to a 49\% improvement in model performance in overall loss, as shown in the second row in Table~\ref{tab:generalization}.

These findings underscore the importance of comprehensive baseline mapping to collect sufficiently diverse training data. This is essential for training models that are robust enough to perform accurate localization during reinspection tasks.

\begin{table}[t]
\centering
\caption{Performance of configuration C6 on dataset D3.  $\mathcal{L}_\mathbf{p}$ and $\mathcal{L}_\theta$ are median values across the test data. $\mathcal{L}$ was calculated using Eq.~\ref{eqn:my_loss_func} with the average distance $d = 3$~m. The best performance for each metric is highlighted in bold.}
\resizebox{\columnwidth}{!}{%
\begin{tabular}{cccccc}
\toprule
\multirow{2}{*}{\textbf{Training Dataset}} & \multirow{2}{*}{\textbf{EKF}} & \multirow{2}{*}{\textbf{Color Jittering}} & \multicolumn{3}{c}{\textbf{Performance Metrics}} \\
\cmidrule(lr){4-6}
& & & $\mathcal{L}$ (m) & $\mathcal{L}_\mathbf{p}$ (m)& $\mathcal{L_\theta}$ (\textdegree) \\
\midrule
D1 & & & 1.45 & 1.34 & 2.09 \\
D1+D2 & & & 0.75 & 0.58 & 3.20 \\
D1+D2 & \cmark & & 0.47 & 0.47 & 0.00 \\
D1+D2+D4 & & & 0.52 & 0.40 & 2.28 \\
D1+D2+D4 & & \cmark & 0.20 & 0.15 & 0.93 \\
D1+D2+D4 & \cmark & \cmark & \textbf{0.11} & \textbf{0.11} & \textbf{0.00}  \\
\bottomrule
\end{tabular}}
\label{tab:generalization}
\end{table}

\begin{figure}[ht]
    \centerline{\includegraphics[width=0.5\textwidth]{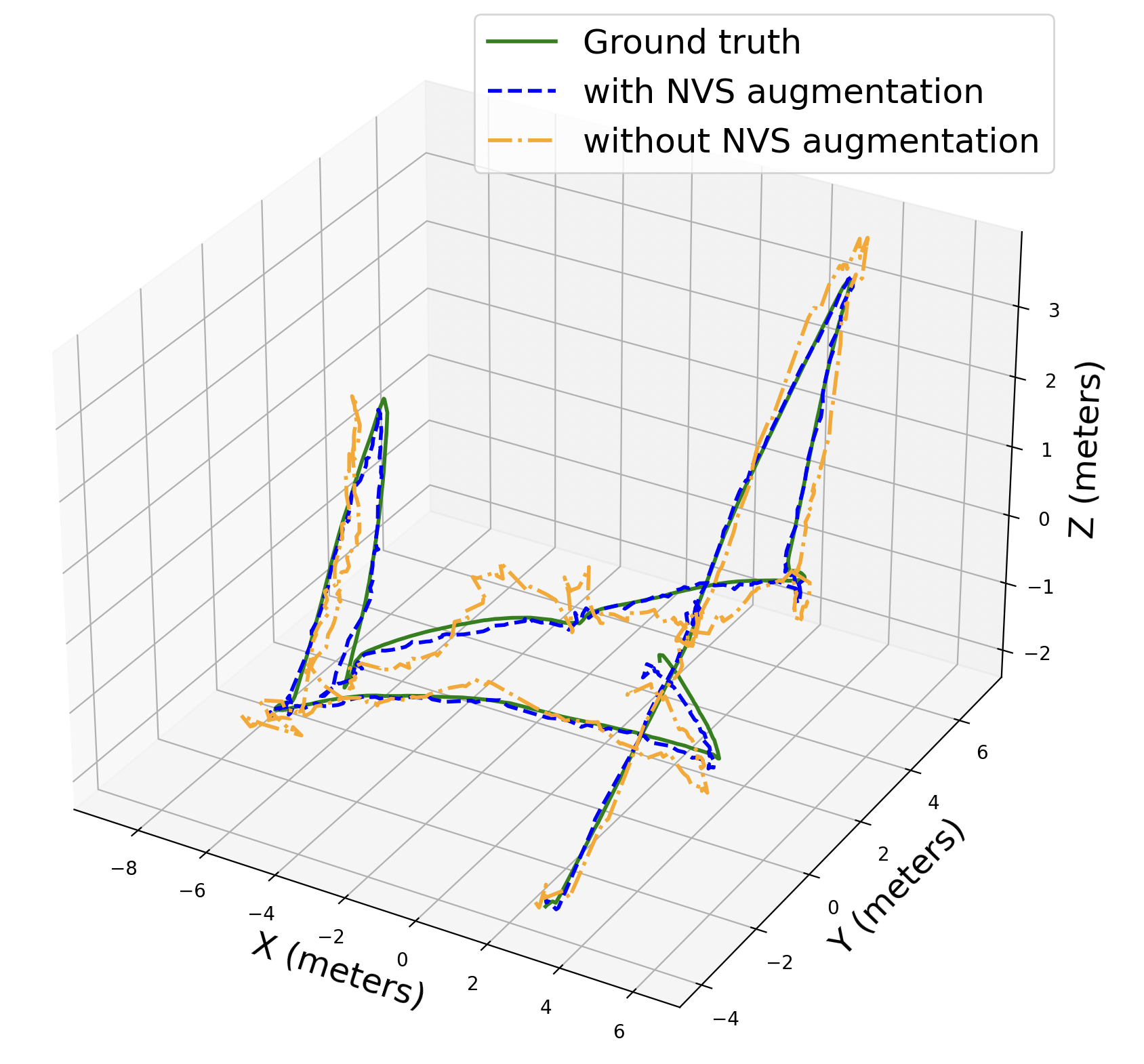}}
\caption{3D trajectory comparison of ground truth versus model estimations with and without NVS augmentation in controlled environment. The model trained with NVS augmentation closely follows the ground truth trajectory, indicating improved positional accuracy over the model trained without NVS augmentation.}
\label{fig:tcoms_nerf_traj}
\end{figure}
\section{Augmented Training with Novel View Synthesis}
\label{sec:3}
The previous section demonstrated the importance of diverse training data with good coverage of the surveyed location. Although it may sometimes be possible to collect such data by extensively covering areas during the baseline mapping run, the practical constraints of cost and labor often limit this approach or render it infeasible. We explore alternative approaches to improve model performance in such data-limited scenarios. We propose to use NVS techniques to create models of the 3D scene, and then use these to generate more images from new aspects to augment the training data. In this section, we present the methods of augmenting training data using NVS models and the results of this approach.

\subsection{Methods}
We first select 540 images from D1 and D2 to train an NVS model for the TCOMS scene. For this, we employ \textit{COLMAP}~\cite{schonberger_pixelwise_2016,schonberger_structure--motion_2016}, an open-source Structure-from-Motion computation software, to compute the camera pose associated with each image within an arbitrary reference coordinate. 
We use \textit{nerfstudio}~\cite{nerfstudio} for training the NVS model and rendering images, using the \textit{nerfacto} approach. To enhance the model's robustness to dynamic elements in the scene such as lighting changes, we employ the robust loss function proposed by Sabour et al~\cite{sabour_robustnerf_2023}. The details of training the model are presented in our previous work~\cite{too_oceans_2024}.

To render images for new poses that were not sampled during the ROV run, we first synthesize these new camera poses based on the existing poses in the training dataset. This is done by varying the depth and distance to the structure in the existing poses, keeping the orientation the same to ensure the camera is pointing towards the structure in the newly generated poses. In total, we generate 4193 images, and we refer to this dataset as D4. We then use D1, D2, and D4 for training, and D3 for validation and testing to test the improvement provided by using the NVS-based augmentation. 

Additionally, it is noted that the images in D4 exhibited different brightness levels and background noise as compared to the original data, introduced during the NVS model reconstruction. To address the potential degradation due to this, we further augment the data by jittering the color of each image during training, thus making the pose estimator robust to minute color and lighting changes. For evaluation, we use the same GPU, framework, and hyperparameter tuning methods as described in the previous section.

\subsection{Results \& Discussion}

Our results show that utilizing augmented training data generated by a NVS model leads to a significant enhancement in localization accuracy. Comparing row 2 and row 4 in Table~\ref{tab:generalization}, we find that by augmenting the training data with D4, the overall localization error can be reduced by 30\%. 

\begin{figure}[t]
    \centering
    \begin{subfigure}[b]{\columnwidth}
        \includegraphics[width=\textwidth]{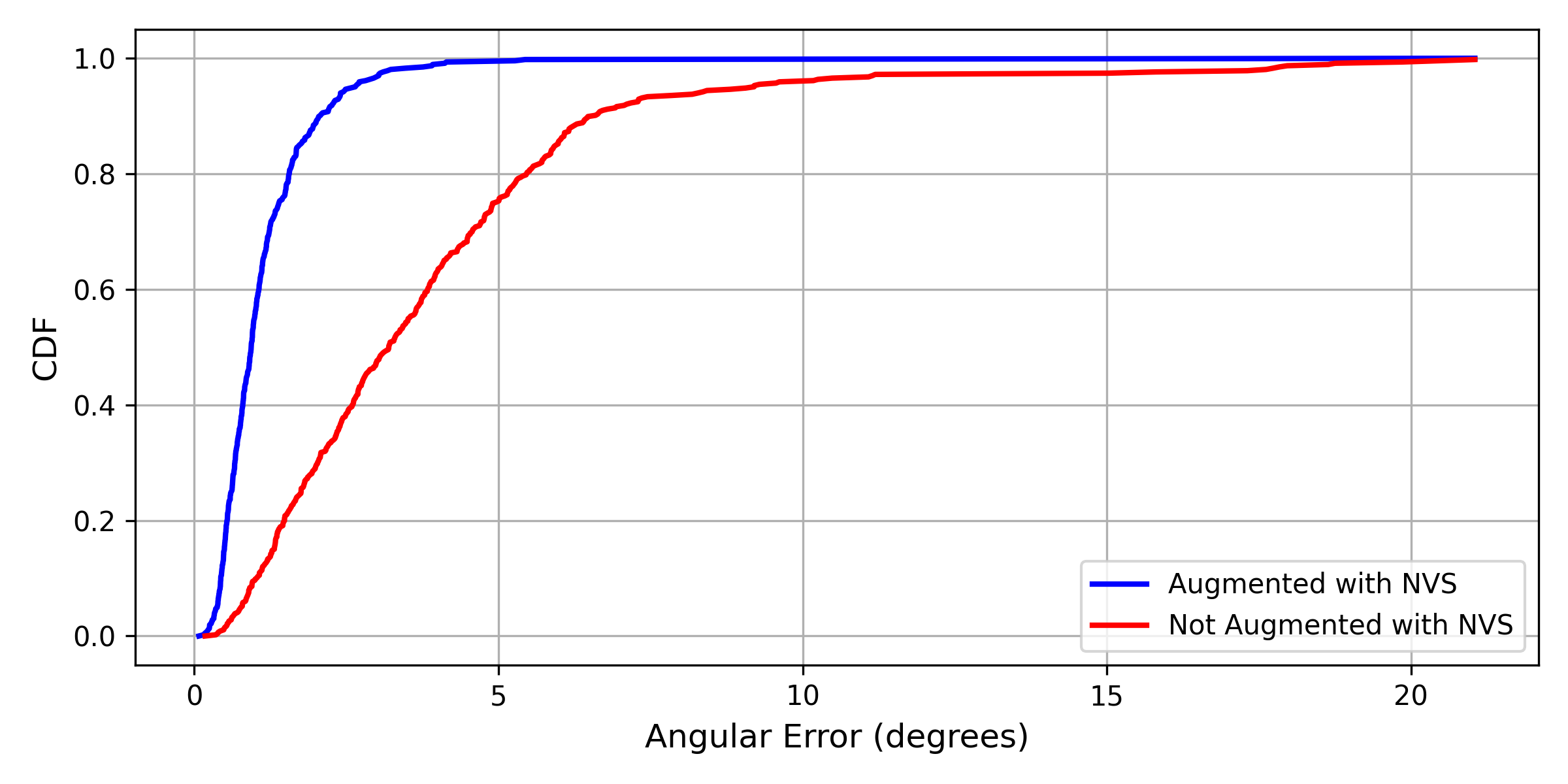}
        \caption{}
        \label{fig:tcoms_nerf_angle}
    \end{subfigure}
    \hfill
    \begin{subfigure}[b]{\columnwidth}
        \includegraphics[width=\textwidth]{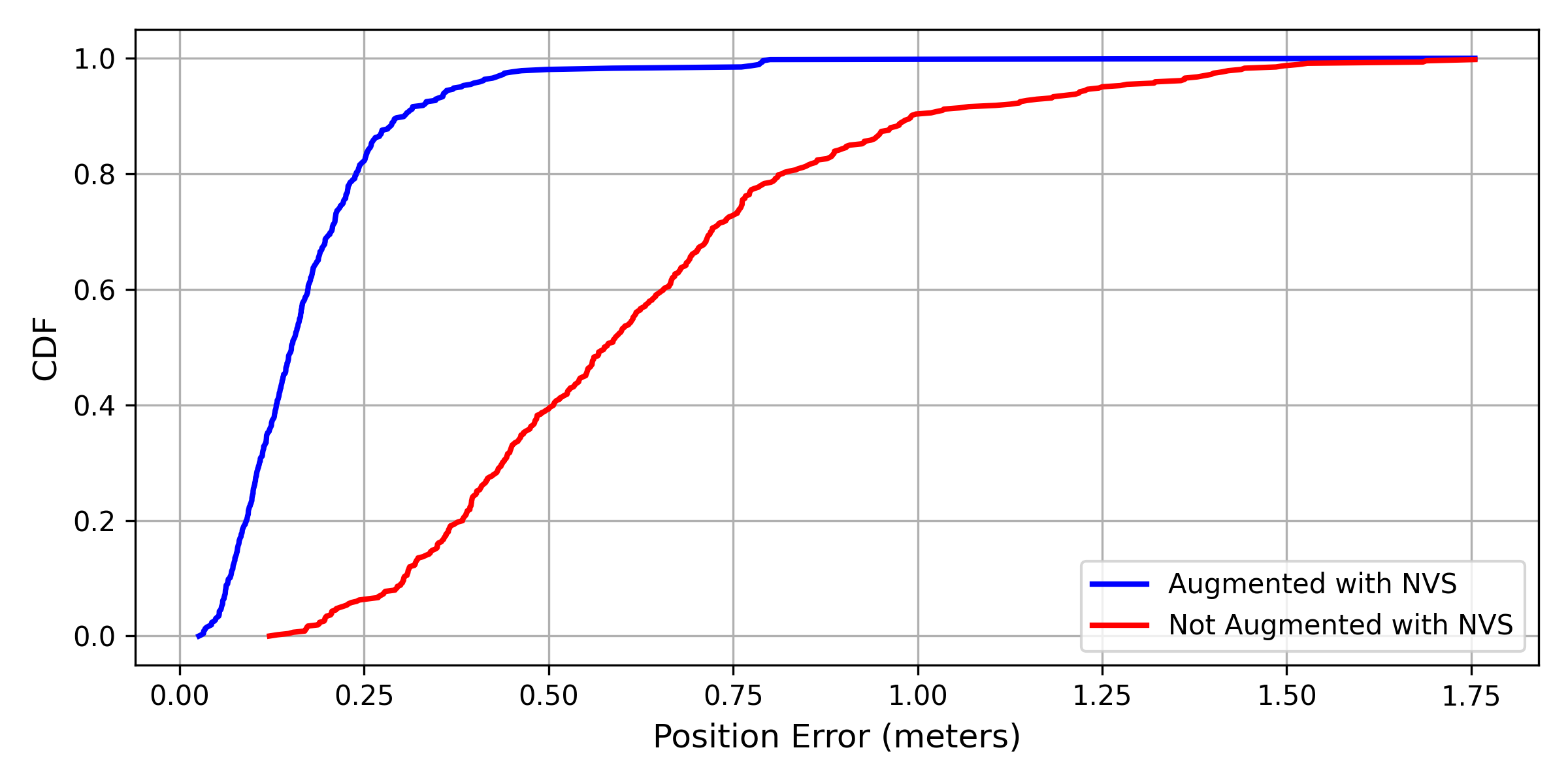}
        \caption{}
        \label{fig:tcoms_nerf_position}
    \end{subfigure}
    \caption{Cumulative distribution function (CDF) of (a) orientation and (b) position errors for models trained with and without NVS augmentation in controlled environment. The plots show that augmented training with NVS yields significantly lower errors for both orientation and position compared to training without augmentation.}
    \label{fig:tcoms_nerf_errors}
\end{figure}

Color jittering augmentation is also highly effective in further improving the model performance, further reducing the error by an additional 61.5\%. We compare the performance of the augmented training with color jittering with the performance without augmented training in Fig.~\ref{fig:tcoms_nerf_traj} and Fig.~\ref{fig:tcoms_nerf_errors}. These plots show that the proposed augmented training with NVS significantly improves the pose estimator's accuracy and reliability in terms of both position and orientation.

\begin{figure}[t!]
    \centering
    \begin{subfigure}{0.45\columnwidth}
        \centering
        \includegraphics[width=\textwidth]{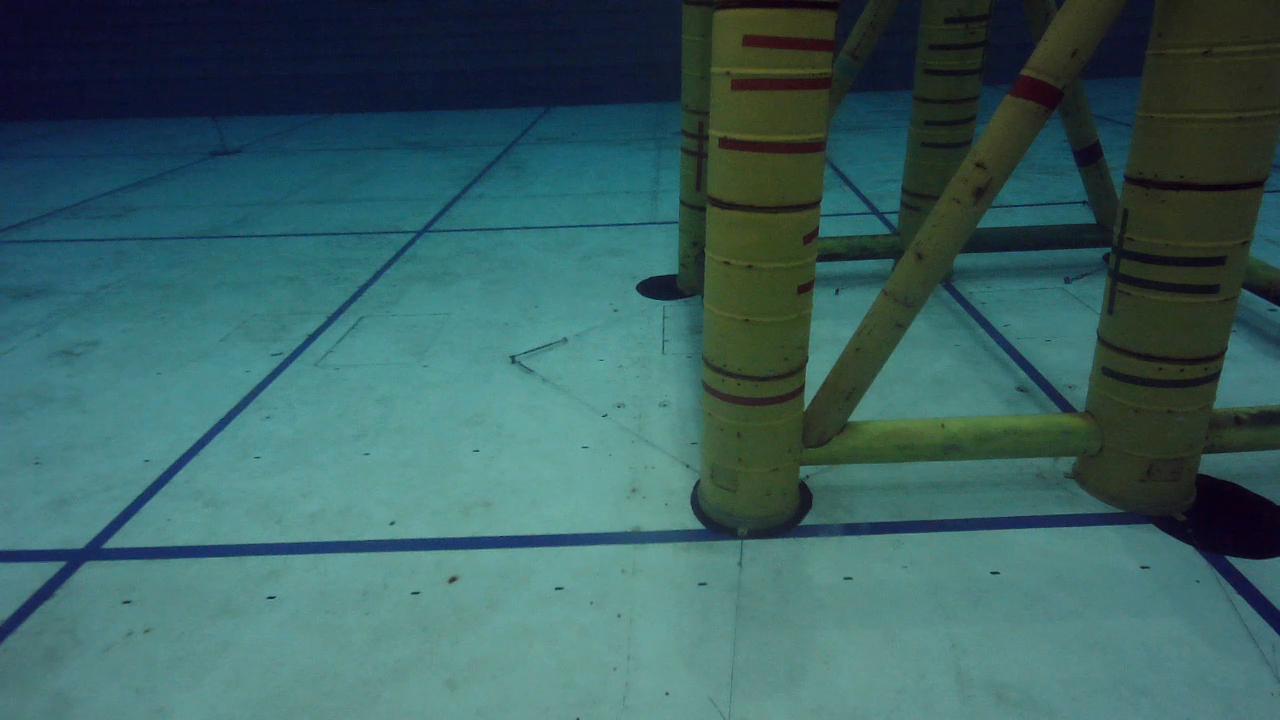}
       \caption{}
    \end{subfigure}\hfill
    \begin{subfigure}{0.45\columnwidth}
      \centering
      \includegraphics[width=\textwidth]{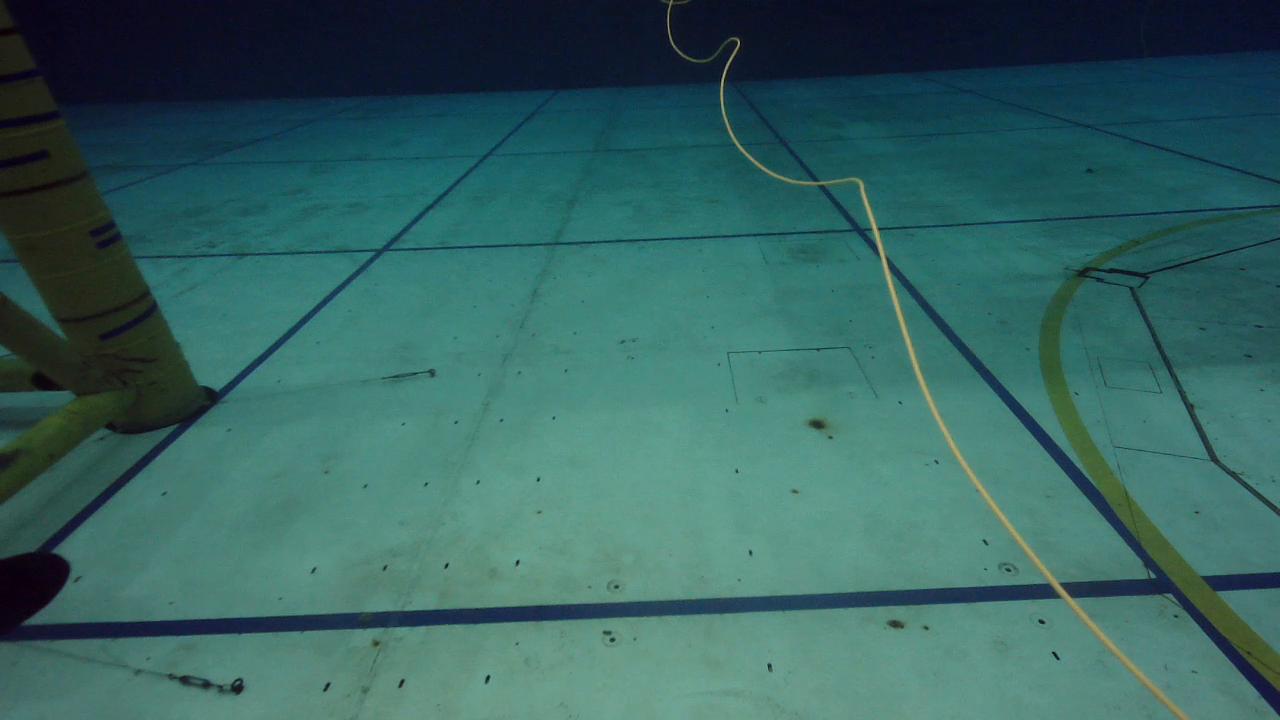}
       \caption{}
    \end{subfigure}\hfill
    \caption{Test images from Clear Water-Mid with amongst the best and worst pose estimation accuracy. Panel (a) is the image with one of the best pose estimation accuracy and panel (b) is the image with one of the worst pose estimation accuracy.}
    \label{fig:tcoms_test_images}
\end{figure}

Nonetheless, we observed the presence of outliers. Upon examining the data, we found that these outliers were caused by transient objects, such as the tether shown in Fig.~\ref{fig:tcoms_test_images}(b), which were not present in the training data.

\section{Localization enhancement via sensor data fusion}
\label{sec:4}
While the trained pose estimators yield small median orientation and position errors, their estimates exhibit some volatility. Our model currently treats each sample independently, ignoring temporal context, and utilizes only the camera inputs during deployment. However, additional information, such as temporal information and other sensor inputs from the ROV, is available. To enhance localization accuracy and achieve a more stable trajectory estimation, we propose sensor fusion using an EKF. This section details the integration of the pose estimator with additional sensor data and presents the results of the sensor fusion.

\subsection{Methods}
Given the sequential nature of data in reinspection missions and the availability of additional sensors, incorporating temporal information and other sensor data presents a viable strategy for improving the model's estimation stability and accuracy. Currently, the visual localization model without sensor fusion occasionally results in estimation of poses that are physically implausible or outliers, in context of the dynamics from previous poses. By integrating knowledge of the ROV's physics model and leveraging previous pose estimates, we can enhance pose accuracy and stability.

Furthermore, during reinspection missions, ROVs are commonly equipped with altimeters and compasses, which have a reasonable accuracy. As such, we could use these reliable depth and orientation measurements during reinspection to further improve the overall localization accuracy. 

\subsubsection{Model}
We assume that the vehicle moves with a constant translational velocity and constant angular velocity since the vehicle normally moves slowly during inspection missions.

We consider the pose estimator outputs, compass data and altimeter data as measurements. The compass yields orientation measurements in the form of a quaternion. The altimeter provides z-coordinate measurements. The pose estimator outputs comprise (1) position in x, y and z coordinates, and (2) orientation in the form of a quaternion. We use an EKF with this model to integrate measurements from the model and these sensor measurements.  

\subsubsection{Measurement noises}
The measurement noise is the other hyperparameter that needs to be carefully selected in the EKF. Nominal values for the measurement noise standard deviations for the compass and altimeter are available from the specifications provided by their manufacturers, and can be set accordingly. However, the noise associated with the pose estimator presents a more complex challenge. Setting a static value for the pose estimator's measurement noise—such as the standard deviation of localization error derived from validation performance—is inadequate. This is due to the inconsistent nature of the network's estimations, which can sometimes exhibit substantial errors. To more accurately represent the dynamic noise in pose estimator, we employ \textit{dropout} techniques at test time for Monte Carlo sampling from the model's posterior distribution.

\textit{Dropout} is a technique commonly used as a regularizer in training neural networks to prevent overfitting. Recent works have shown that using dropout during inference can be used to approximate Bayesian inference over the distribution of the network's weights at test time, without requiring any additional model parameters~\cite{kendall_modelling_2016}. We sample 100 Monte Carlo dropout realizations for the inference of each image input sample, and use the standard deviation of these dropout samples as the measurement noise for the inference sample.

\subsection{Results \& Discussion}
\begin{figure}[t]
    \centerline{\includegraphics[width=0.5\textwidth]{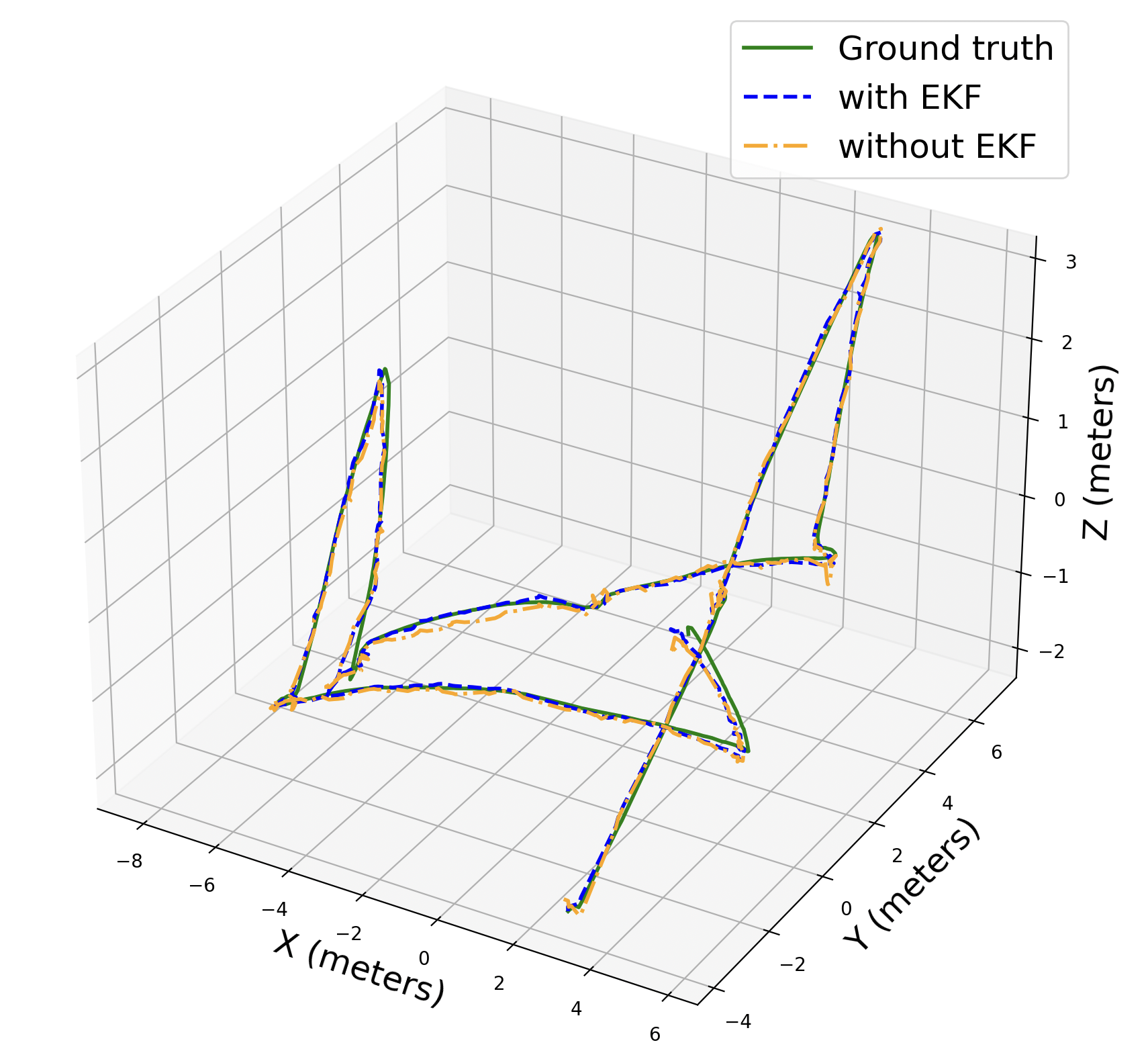}}
\caption{3D trajectory comparison of ground truth versus model estimates with and without EKF. The EKF output closely follows the ground truth trajectory, indicating improved positional accuracy over the pose estimator output.}
\label{fig:tcoms_ekf_traj}
\end{figure}

We tune the process noise standard deviation as a hyperparameter to get the best performance. As shown in Table.~\ref{tab:generalization}, sensor fusion with the EKF yields more accurate estimated poses. We also find that the predicted trajectory using the EKF is smoother compared to that without using the EKF, as illustrated in Fig.~\ref{fig:tcoms_ekf_traj}. 
However, the inference time using the EKF is about ten times as long. 

\section{Field trials at sea}
\label{sec:5}
To further validate our proposed methods, we conducted field trials in a bay near St. John's Island, Singapore (SJI). In this section, we present the methods, results and challenges encountered in using our proposed methods from the previous section in a real-world setting.

\subsection{Methods}
We used the ROV to collect data in an at-sea environment, inspecting a submerged pillar. The pillar selected was approximately 5~m tall and 0.5~m in diameter. Although the pillar was a simple black metallic structure, the barnacles and algae growing on its surface provided visual features that could be used for pose estimation. We drove the ROV following a vertical lawnmower path around the pillar, while recording the video from the camera. Due to the high turbidity in the water, we operated the ROV in close proximity to the structure with the average distance being 1~m. 

\begin{figure*}[t!]
    \centering
    \begin{subfigure}{0.45\columnwidth}
        \centering
        \includegraphics[width=\textwidth]{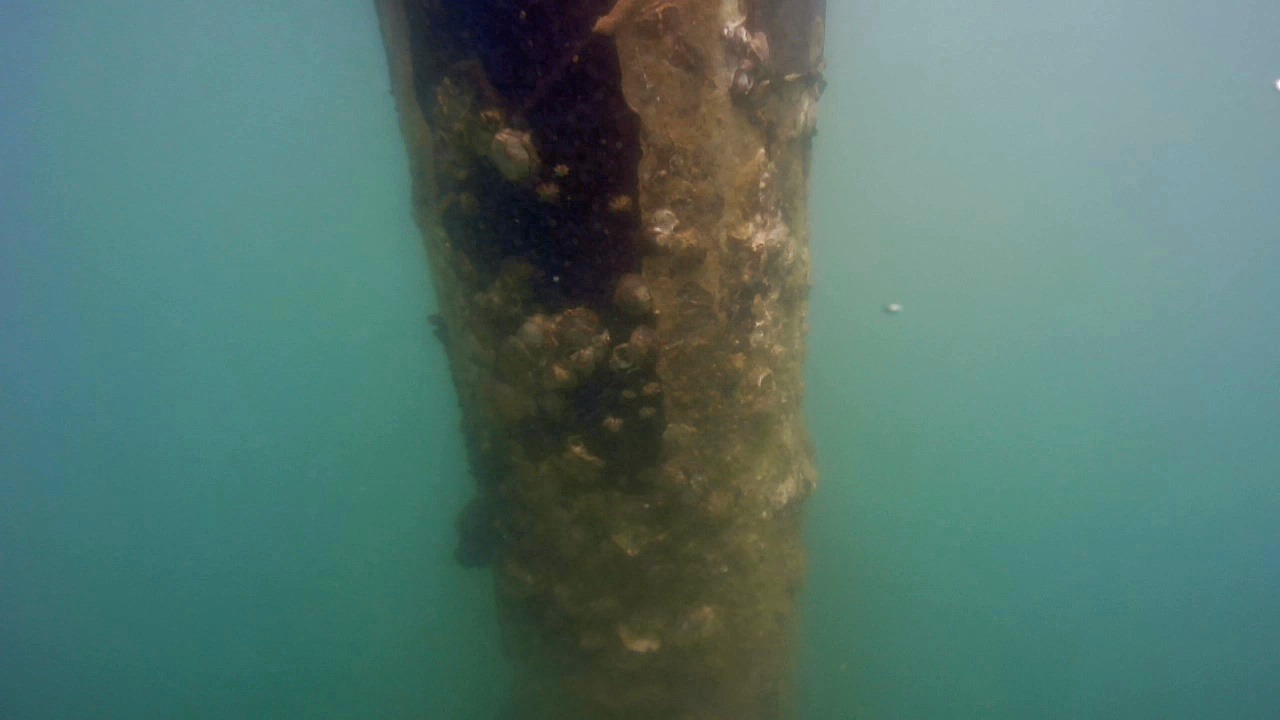}
       \caption{}
    \end{subfigure}\hfill
    \begin{subfigure}{0.45\columnwidth}
        \centering
        \includegraphics[width=\textwidth]{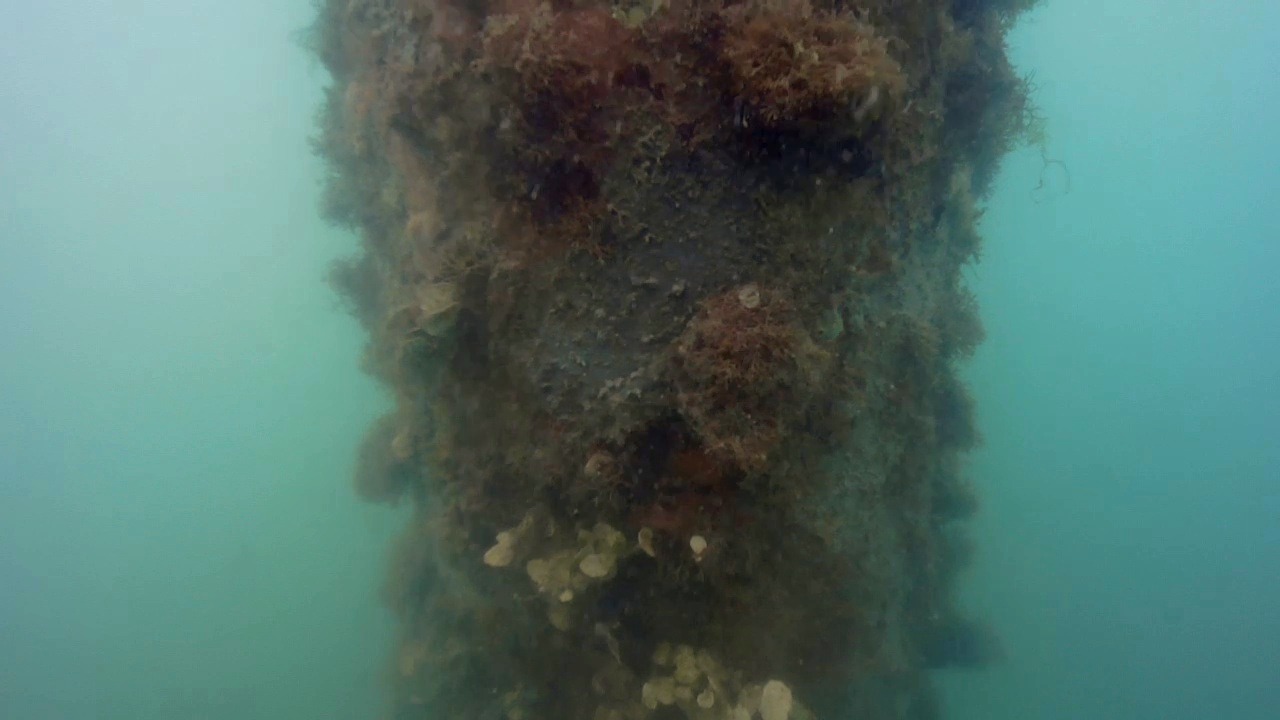}
       \caption{}
    \end{subfigure}\hfill
    \begin{subfigure}{0.45\columnwidth}
      \centering
      \includegraphics[width=\textwidth]{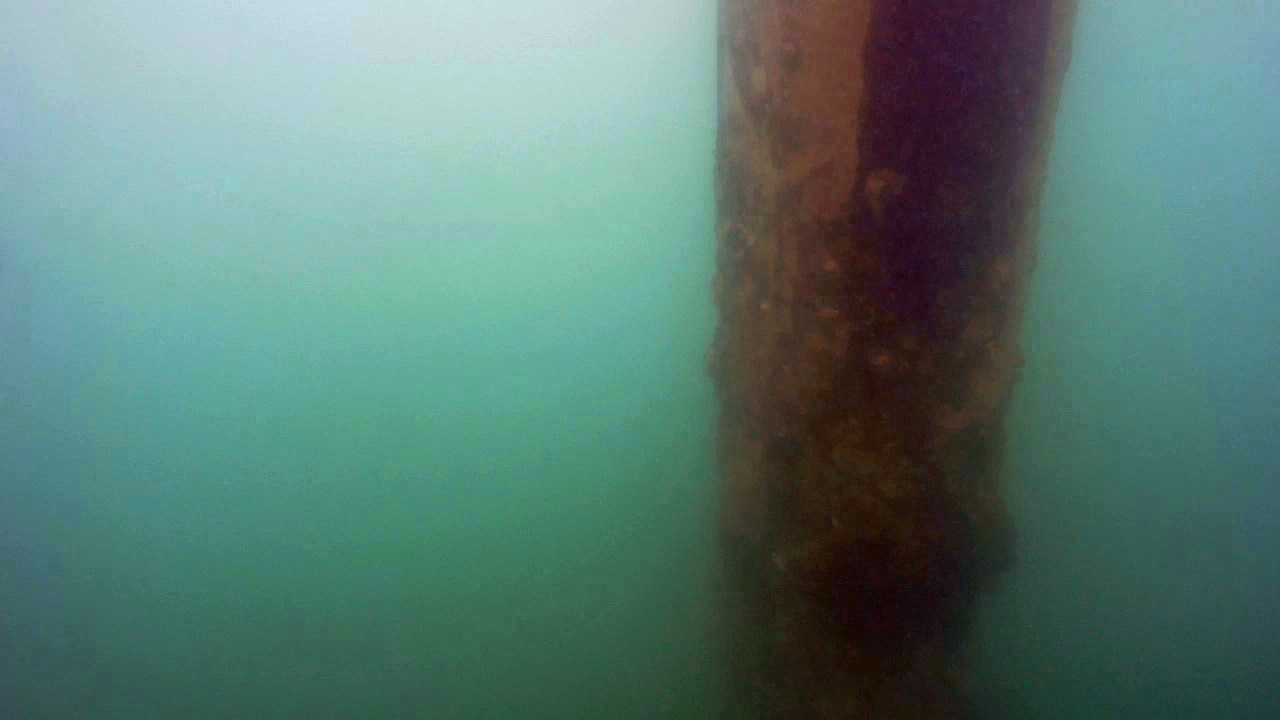}
       \caption{}
    \end{subfigure}\hfill
    \begin{subfigure}{0.45\columnwidth}
      \centering
      \includegraphics[width=\textwidth]{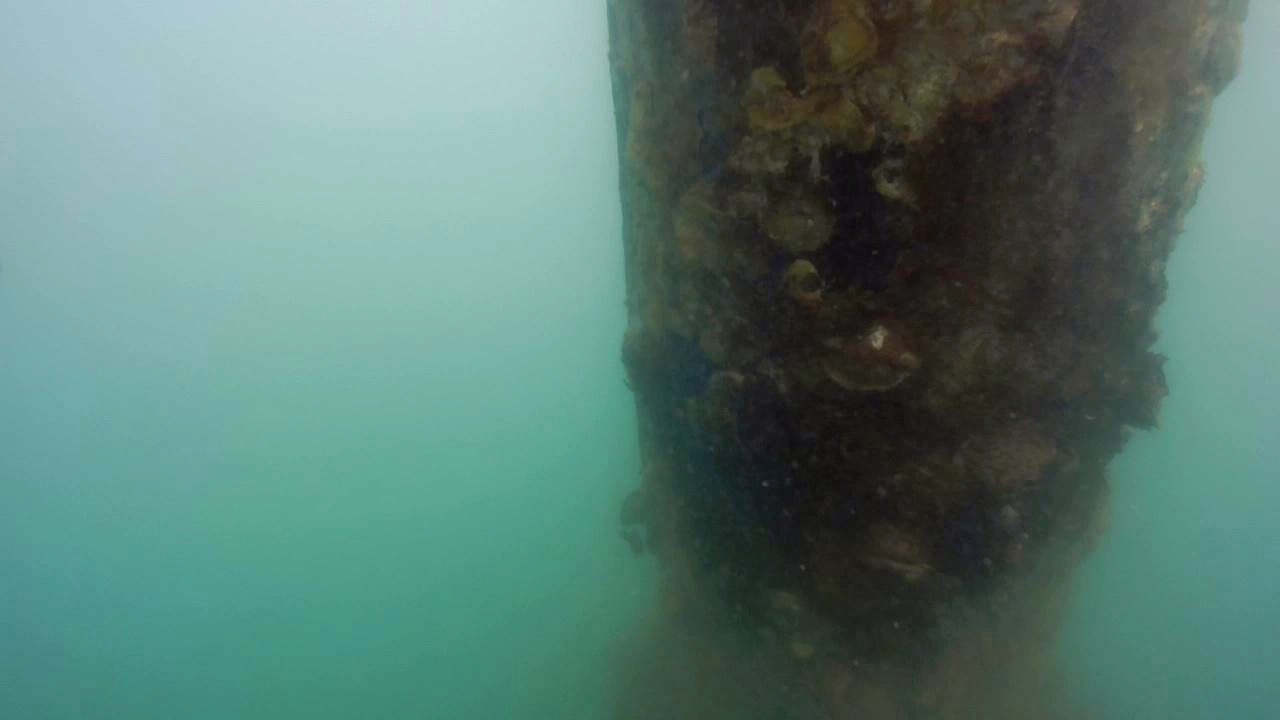}
        \caption{}
    \end{subfigure}
    
    \caption{Sample images from Sea Water-1 and Sea Water-2 dataset. Panels (a) and (b) are from Sea Water-1 dataset, and (c) and (d) are from Sea Water-2 dataset. The images from Sea Water-2 dataset show higher turbidity and thus fewer features than images from Sea Water-1.}
    \label{fig:sji_cam_images}
\end{figure*}

We collected two datasets, named as D5 and D6, on two different days. Samples of images collected in these datasets are shown in Fig.~\ref{fig:sji_cam_images}.

\begin{figure}[t]
    \centering
    \begin{subfigure}{0.45\columnwidth}
        \centering
        \includegraphics[width=\textwidth]{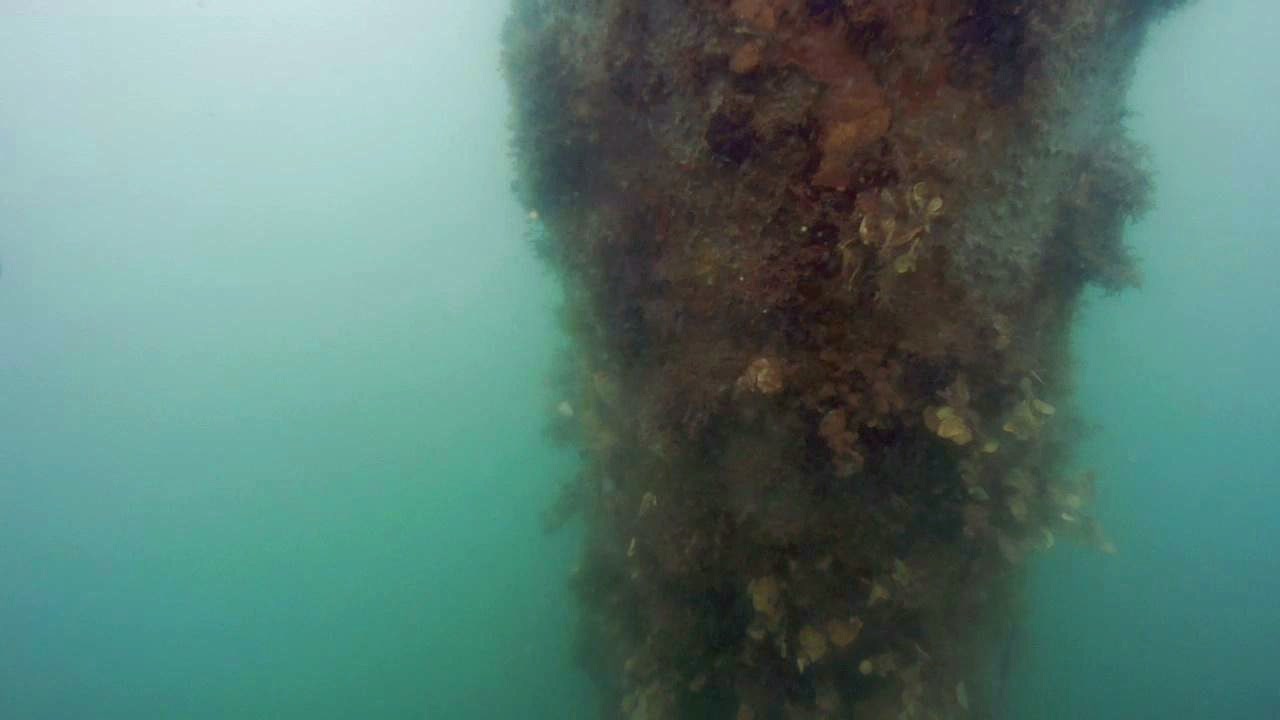}
       \caption{}
    \end{subfigure}\hfill
    \begin{subfigure}{0.45\columnwidth}
      \centering
      \includegraphics[width=\textwidth]{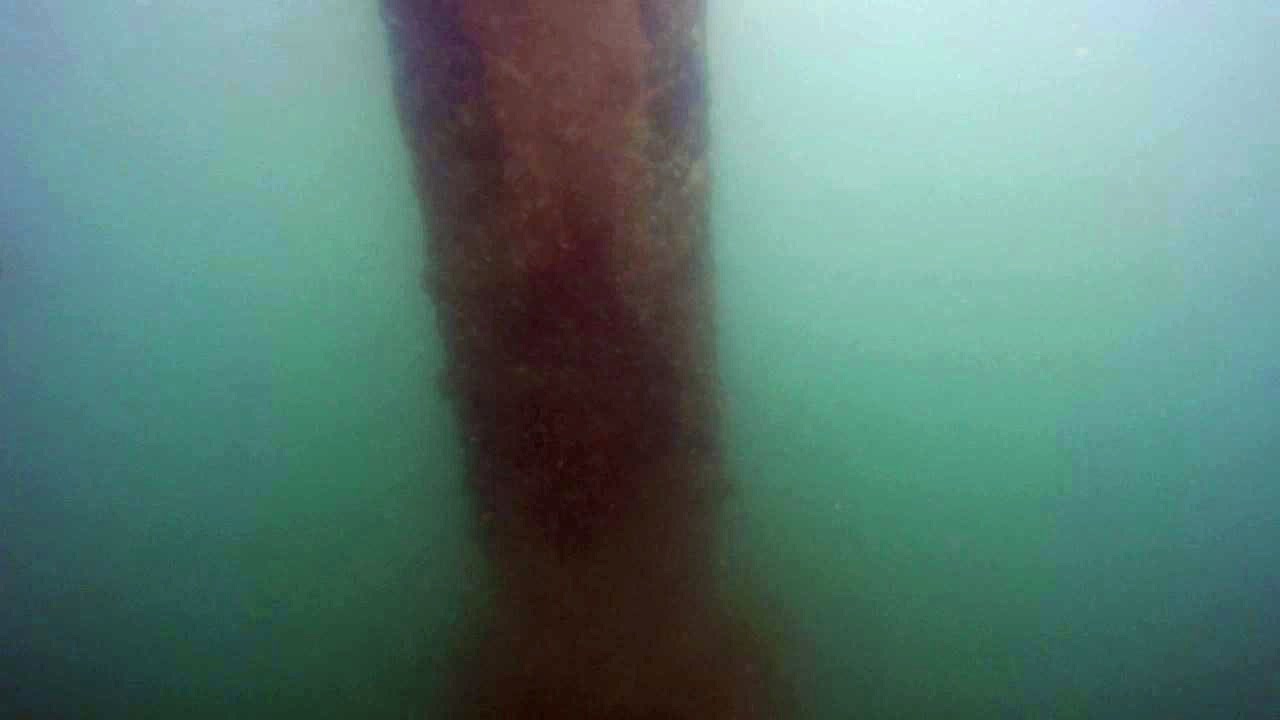}
      \caption{}
      \label{fig:sji_bad_pose}
    \end{subfigure}\hfill
    \caption{Test images from Sea Water-2 with amongst the best and worst pose estimation accuracy. Panel (a) is the image with one of the best pose estimation accuracy and (b) is the image with one of the worst pose estimation accuracy.}
    \label{fig:sji_test_images}
\end{figure}

\begin{figure*}[t]
    \centering
    \makebox[0.24\textwidth]{\textbf{Camera Image}} 
    \hfill
    \makebox[0.72\textwidth]{\textbf{Rendered Images}}
    \hfill
    \vspace{5pt} 

    \fbox{ 
    \begin{subfigure}[t]{0.224\textwidth}
        \includegraphics[width=\textwidth]{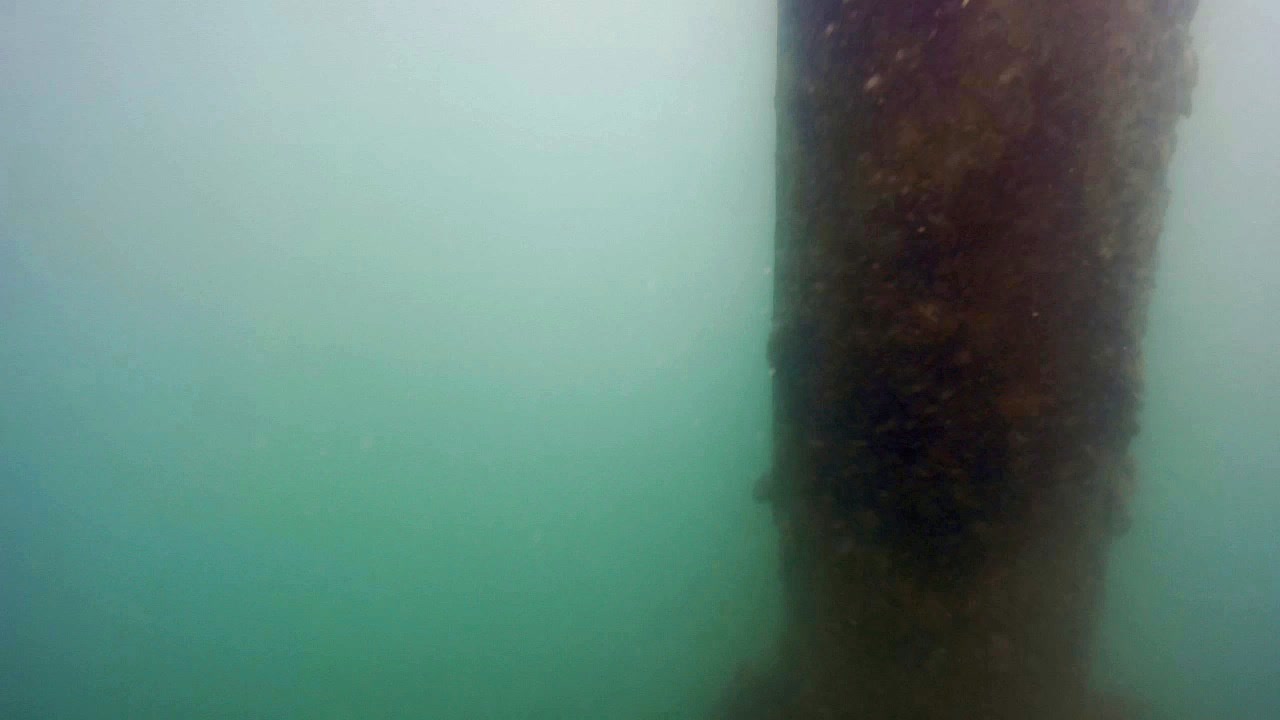}
    \end{subfigure}
    }
    \hfill
    \fbox{ 
    \begin{minipage}[t]{0.7\textwidth}
        \begin{subfigure}[t]{0.32\textwidth}
            \includegraphics[width=\textwidth]{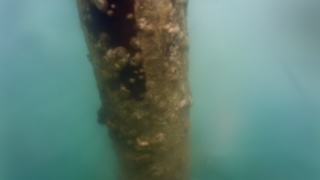}
        \end{subfigure}
        \hfill
        \begin{subfigure}[t]{0.32\textwidth}
            \includegraphics[width=\textwidth]{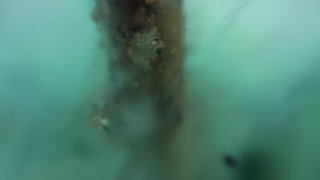}
        \end{subfigure}
        \hfill
        \begin{subfigure}[t]{0.32\textwidth}
            \includegraphics[width=\textwidth]{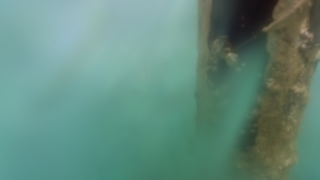}
        \end{subfigure}
    \end{minipage}
    }
    \caption{Camera image and NVS rendered images in the bay near SJI. The rendered images produce photorealistic views of the structure but exhibit some artifacts and noise depending on the camera pose.}
    \label{fig:sji-images}
\end{figure*}

We use D5 to train an NVS model following the method described in Section~\hyperref[sec:2]{II}. New camera poses are generated using the same approach. The NVS model is then utilized to create an augmented training dataset, named D7. Samples of images generated at new poses using the NVS model are shown in Fig.~\ref{fig:sji-images}.

We train the best visual localization architecture configuration, C6, both with augmented training data (datasets D5+D7) and without any augmentation (only D5). Dataset D6 is used for validation and testing. The training methods are similar to those described in Section~\hyperref[sec:2]{II}.

\subsection{Results \& Discussion}
\begin{figure}[h]
    \centering
    \begin{subfigure}[b]{\columnwidth}
        \includegraphics[width=\textwidth]{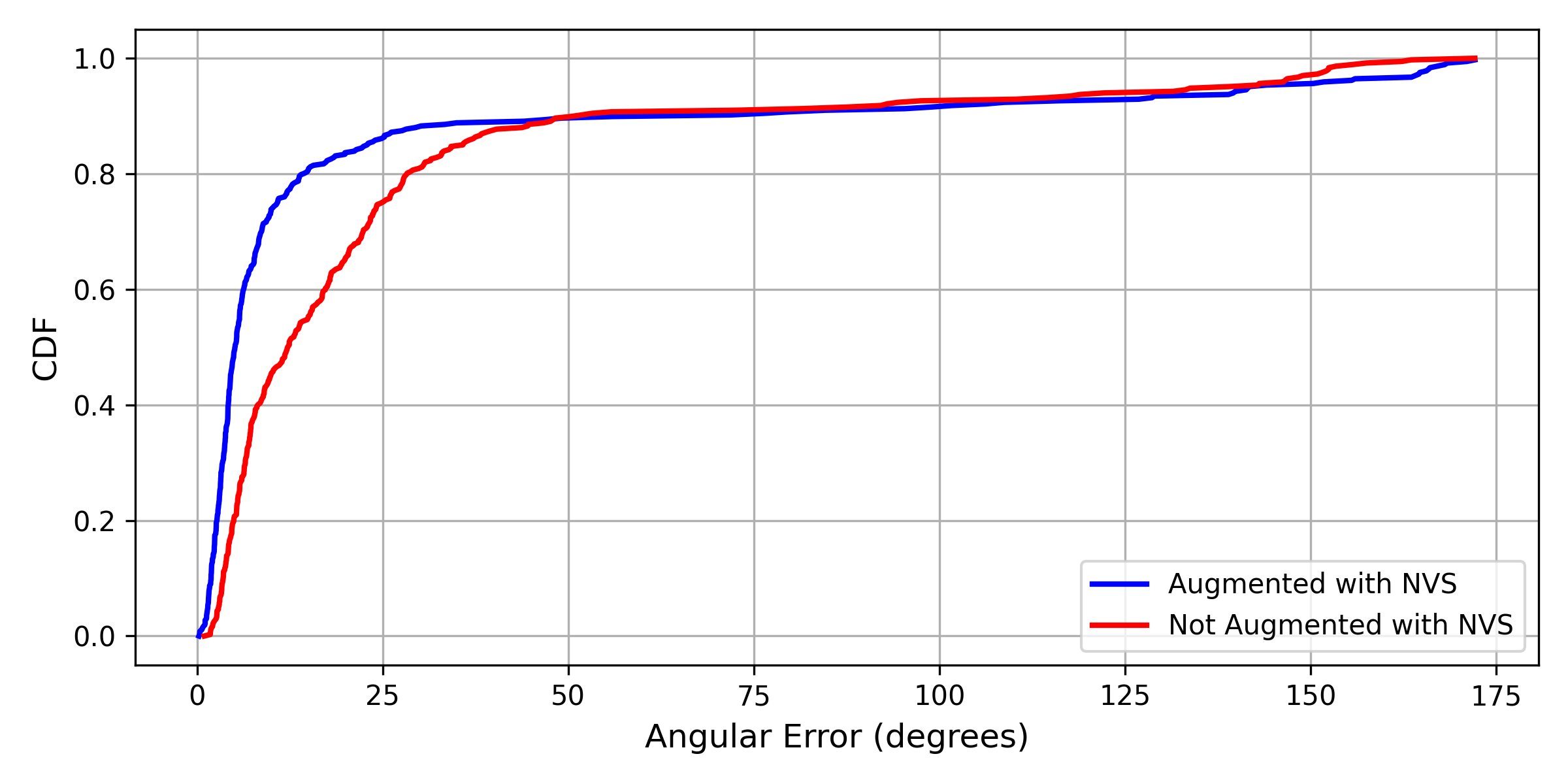}
        \caption{}
        \label{fig:sji_nerf_angle}
    \end{subfigure}
    \hfill
    \begin{subfigure}[b]{\columnwidth}
        \includegraphics[width=\textwidth]{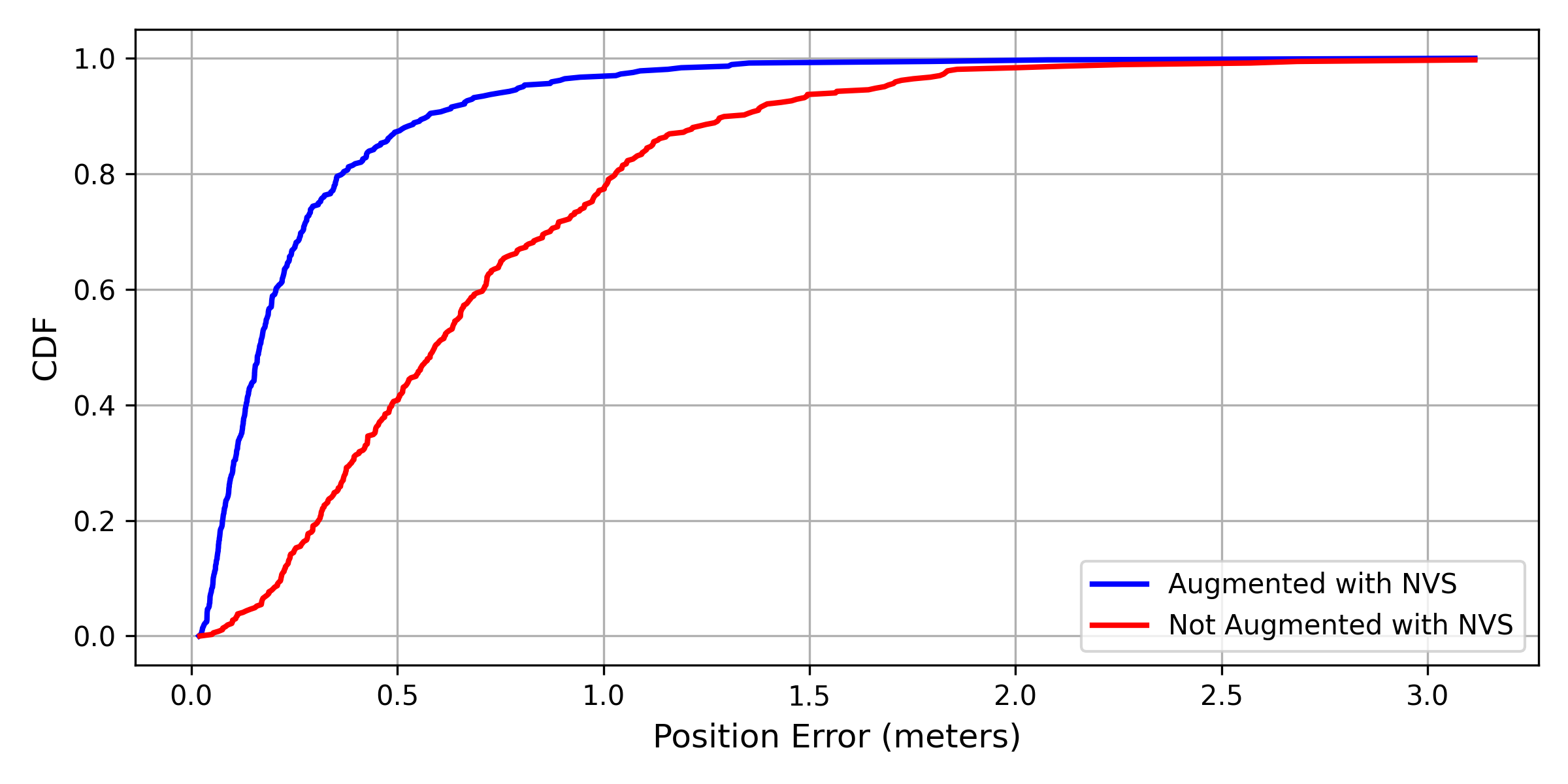}
        \caption{}
        \label{fig:sji_nerf_position}
    \end{subfigure}
    \caption{CDF of (a) orientation and (b) position errors for models trained with and without NVS augmentation in at-sea environment. The plots show that augmented training with NVS yields significantly lower errors for both orientation and position compared to training without augmentation.}
    \label{fig:sji_nerf_errors}
\end{figure}

As shown in Fig.~\ref{fig:sji_nerf_errors}, augmented training with NVS yields significant improvement in both position and orientation accuracy compared to training without NVS augmentation. With configuration C6 and augmented training, we are able to achieve a position accuracy of 0.17~m and orientation accuracy of 5.09\textdegree. We present the performance of C6 on D6 in Table~\ref{tab:sea}. While the median accuracy is comparable to the performance in the controlled environment, we note that the standard deviation in the errors are much larger at sea.  

Clearly, the real-world setting at sea presents several challenges that are not present in controlled environments. The biggest challenge is the turbidity of the water, which significantly affects the quality of the images. Moreover, lighting is inconsistent at different camera poses and on different days, causing high variablity in the image quality. This introduced three new challenges. First, the noisy images make it challenging to compute camera poses in \textit{COLMAP}, resulting in a sparse number of registered images. Consequently, the EKF model could not be used for performance improvement since it would not be feasible to assume constant velocity and angular velocity in the vehicle model. Second, the turbidity and inconsistent lighting in the training data introduced artifacts in the NVS model. Thus, the rendered images are more noisy compared to images in clear waters, as shown in Fig.~\ref{fig:sji-images}. Third, the high variability in image quality can lead to more estimation outliers and large errors during inference. All of these contribute to a decrease in the model's performance. Nevertheless, we note that the NVS model is still able to produce photorealistic views of the structure.

\begin{table}[t]
    \centering
    \caption{Performance of configuration C6 on dataset D6.  $\mathcal{L}_\mathbf{p}$ and $\mathcal{L}_\theta$ are median values across the test data. $\mathcal{L}$ was calculated using Eq.~\ref{eqn:my_loss_func} with the average distance $d = 1$~m.}%
    \resizebox{\columnwidth}{!}{%
    \begin{tabular}{ccccc}
    \toprule
    \multirow{2}{*}{\textbf{Training Dataset}} & \multirow{2}{*}{\textbf{Color Jittering}} & \multicolumn{3}{c}{\textbf{Performance Metrics}} \\
    \cmidrule(lr){3-5}
    & & $\mathcal{L}$ (m) & $\mathcal{L}_\mathbf{p}$  (m)& $\mathcal{L_\theta}$ (\textdegree )\\
    \midrule
    D5 &  & 0.80 & 0.59 & 12.15 \\
    D5+D7 & \cmark & \textbf{0.26}& \textbf{0.17}& \textbf{5.09}\\
    \bottomrule
    \end{tabular}}
    \label{tab:sea}
\end{table}

\section{Conclusion}
\label{sec:6}
In this paper, we addressed the challenge of localization in underwater inspection missions with a neural-network based pose estimator. We proposed a new loss function to train the pose estimator, and demonstrated that training with $d$-loss significantly improved the model's performance in pose estimation tasks. This improvement is attributed to the incorporation of domain-specific physics, as the $d$-loss accounts for the relevant geometric considerations in the inspection mission. Furthermore, this loss function also lends more interpretability to the loss. Employing the ResNet50 backbone with a CNN+LSTM architecture allows us to efficiently use the available visual information to estimate the pose, and yielded improvements in the localization performance as compared to benchmark architectures. 

In terms of the generalization, using more diverse data with a wider distribution significantly enhances the localization performance on test data that lies outside the training distribution. We additionally investigated the use of NVS techniques to augment training data and showed that this significantly improves the estimator's performance with previously unsurveyed poses. Thus, this provides a cost-effective and information-efficient method to improve the generalization performance without having to undertake expensive field trials to collect additional data. Further integrating the pose estimator with an EKF allows us to fuse sensor data with the visual-based estimates, and we demonstrated that this further improved the performance and stability. We validated our proposed methods in both controlled environments in a clear water tank and real-world settings at sea.

Overall, our results show that our proposed methods significantly improve the visual localization performance in both controlled underwater environments and real-world settings and achieve good localization accuracy to within desired limits, providing a cost-effective alternative or complement to existing localization solutions. Real-world challenges such as turbidity and noise limit the performance achievable, but the proposed method still performs reasonably, especially when data augmentation using color-based augmentation is used to robustify the technique against color distortion.

Potential improvements to this technique may include utilizing temporal information (i.e., more than one image at a time) to improve the accuracy of pose estimates, fusing more data such as control input information and sonar data, exploring better sensor fusion techniques such as particle filters and using per-pixel loss together with NVS rendered images to fine-tune the model.

The algorithm developed in this work is also utilized as part of a model-based image compression technique for low bandwidth scenarios. The details of this approach and the preliminary results are presented in our previous work~\cite{rajat_oceans_2024}.


%



\section*{Acknowledgment}
This research project is supported by A*STAR under its RIE2020 Advanced Manufacturing and Engineering (AME) Industry Alignment Fund - Pre-Positioning (IAF-PP) Grant No. A20H8a0241.

\ifCLASSOPTIONcaptionsoff
  \newpage
\fi



%
\bibliographystyle{IEEEtran}
\bibliography{references}

%








\end{document}